\documentclass[10pt,journal,compsoc]{IEEEtran}

\usepackage[T1]{fontenc}
\usepackage{amsfonts,amssymb,amsmath}
\usepackage{utfsym}
\usepackage{bm}
\usepackage{balance}
\usepackage{algorithm}
\usepackage{algorithmicx}
\usepackage[noend]{algpseudocode}
\usepackage{makecell}
\usepackage{graphicx}
\usepackage{textcomp}
\usepackage{multirow}
\usepackage{booktabs}
\usepackage{float}
\usepackage{stfloats}
\usepackage{url}
\usepackage{ragged2e}
\usepackage{graphics}
\usepackage{subfigure}
\usepackage{colortbl}
\usepackage{setspace}
\usepackage{array}
\usepackage{soul}
\usepackage{hyperref}
\soulregister\cite7
\soulregister{\ref}7
\ifCLASSINFOpdf
\else
\fi
\ifCLASSOPTIONcompsoc
\usepackage[nocompress]{cite}
\else
\usepackage{cite}
\fi
\begin{document}
%
% paper title
% Titles are generally capitalized except for words such as a, an, and, as,
% at, but, by, for, in, nor, of, on, or, the, to and up, which are usually
% not capitalized unless they are the first or last word of the title.
% Linebreaks \\ can be used within to get better formatting as desired.
% Do not put math or special symbols in the title.
%\title{Semi-supervised Anomaly Detection with Limited Annotations: A Feature Encoding Based Approach}
\title{DTAAD: Dual Tcn-Attention Networks for Anomaly Detection in Multivariate Time Series Data}
%
% author names and IEEE memberships
% note positions of commas and nonbreaking spaces ( ~ ) LaTeX will not break
% a structure at a ~ so this keeps an author's name from being broken across
% two lines.
% use \thanks{} to gain access to the first footnote area
% a separate \thanks must be used for each paragraph as LaTeX2e's \thanks
% was not built to handle multiple paragraphs
%

\author{Ling-rui Yu
\IEEEcompsocitemizethanks{\IEEEcompsocthanksitem This paper has been accepted to Knowledge-Based Systems}
}

% The paper headers
%\markboth{Journal of \LaTeX\ Class Files,~Vol.~14, No.~8, August~2015}%
%\markboth{IEEE TRANSACTIONS ON NEURAL NETWORKS AND LEARNING SYSTEMS}%
%{Shell \MakeLowercase{\textit{et al.}}: Bare Demo of IEEEtran.cls for IEEE Journals}
% The only time the second header will appear is for the odd numbered pages
% after the title page when using the twoside option.
% 
% *** Note that you probably will NOT want to include the author's ***
% *** name in the headers of peer review papers.                   ***
% You can use \ifCLASSOPTIONpeerreview for conditional compilation here if
% you desire.

% If you want to put a publisher's ID mark on the page you can do it like
% this:
%\IEEEpubid{0000--0000/00\$00.00~\copyright~2015 IEEE}
% Remember, if you use this you must call \IEEEpubidadjcol in the second
% column for its text to clear the IEEEpubid mark.

% use for special paper notices
%\IEEEspecialpapernotice{(Invited Paper)}

% make the title area
%\maketitle

% As a general rule, do not put math, special symbols or citations
% in the abstract or keywords.
\IEEEtitleabstractindextext{
\begin{justify}
\begin{abstract}
Anomaly detection techniques enable effective anomaly detection and diagnosis in multi-variate time series data, which are of major significance for today’s industrial applications. However, establishing an anomaly detection system that can be rapidly and accurately located is a challenging problem due to the lack of anomaly labels, the high dimensional complexity of the data, memory bottlenecks in actual hardware, and the need for fast reasoning. In this paper, we propose an anomaly detection and diagnosis model, DTAAD, based on Transformer and Dual Temporal Convolutional Network (TCN). Our overall model is an integrated design in which an autoregressive model (AR) combines with an autoencoder (AE) structure. Scaling methods and feedback mechanisms are introduced to improve prediction accuracy and expand correlation differences. Constructed by us, the Dual TCN-Attention Network (DTA) uses only a single layer of Transformer encoder in our baseline experiment, belonging to an ultra-lightweight model. Our extensive experiments on seven public datasets validate that DTAAD exceeds the majority of currently advanced baseline methods in both detection and diagnostic performance. Specifically, DTAAD improved F1 scores by $\boldsymbol{8.38\%}$ and reduced training time by $\boldsymbol{99\%}$ compared to the baseline. The code and training scripts are publicly available on GitHub at \url{https://github.com/Yu-Lingrui/DTAAD}.
\end{abstract}
\end{justify}

% Note that keywords are not normally used for peerreview papers.
}

% For peer review papers, you can put extra information on the cover
% page as needed:
% \ifCLASSOPTIONpeerreview
% \begin{center} \bfseries EDICS Category: 3-BBND \end{center}
% \fi
%
% For peerreview papers, this IEEEtran command inserts a page break and
% creates the second title. It will be ignored for other modes.
\maketitle
\IEEEdisplaynontitleabstractindextext

\IEEEpeerreviewmaketitle

\section{Introduction}
% The very first letter is a 2 line initial drop letter followed
% by the rest of the first word in caps.
% 
% form to use if the first word consists of a single letter:
% \IEEEPARstart{A}{demo} file is ....
% 
% form to use if you need the single drop letter followed by
% normal text (unknown if ever used by the IEEE):
% \IEEEPARstart{A}{}demo file is ....
% 
% Some journals put the first two words in caps:
% \IEEEPARstart{T}{his demo} file is ....
% 
% Here we have the typical use of a "T" for an initial drop letter
% and "HIS" in caps to complete the first word.
\IEEEPARstart{A}{nomaly} detection refers to the detection of these anomalies from the expected data that deviate substantially from the rest of the distribution, and they are then labeled as anomalies. Multivariate time series are ubiquitous in the real world, and anomaly detection in this regard has always been a research hotspot of the times \cite{gupta2013outlier}. Anomaly detection techniques are now widely used in various application domains such as finance \cite{wu2013dynamic}, epidemic \cite{wu2020deep}, industry, cyber hacking, credit fraud, power industry, unmanned vehicles, healthcare, and many data-driven industries like distributed computing, urban IoT, robotics, traffic, and sensor network monitoring \cite{roberts2013gaussian}, satellite aviation, \textit{etc}., and are particularly prominent in data mining, probabilistic statistics~\cite{benkabou2021local}, machine learning \cite{chen2019unsupervised, li2020anomaly}, computer vision \cite{luo2019video}, \textit{etc}. In recent years, anomaly detection methods suitable for Deep Learning (DL) have been all the rage. DL has shown great ability in learning complex data such as temporal data, streaming media data, graphical spatial data, and high-dimensional data, raising the upper limit of various learning tasks. DL for anomaly detection is referred to as deep anomaly detection \cite{bulusu2020anomalous, pang2021deep}, which is performed by deep neural networks to learn features to determine or get anomaly scores to achieve the purpose of anomaly detection, and it is usually performed in the context of prediction.
% \hl{When faced with the deficiency of labeled data, supervised methods often suffer from the problems of performance degradation.} 
% Challenges - lack of labels, diagnosis/root cause analysis, volatile data, quick inference in time-critical systems, low data availability in federated setups

\textbf{Challenges.} In contemporary data science research, detecting anomalies in large-scale databases has become increasingly challenging due to the growing richness and complexity of data types. The proliferation of IoT technology has led to the generation of vast amounts of data from numerous sensors and devices, which pose difficulties not only in volume but also in diversity, complicating accurate data analysis. Moreover, modern applications demand high responsiveness, necessitating rapid reasoning for quick recovery and optimal service quality. Time series databases, often comprising data from various engineering devices interacting with the environment, humans, or other systems, frequently exhibit uncertainty and temporal trends. Hence, accurately identifying anomalies and distinguishing non-conforming observations against the temporal trend is crucial. However, the scarcity of labeled data and the diversity of outlier types exacerbate the challenge. Traditional supervised learning models face limitations in direct application, while unsupervised learning may prove effective in other domains \cite{chalapathy2019deep}. Furthermore, beyond anomaly detection, pinpointing the root cause of anomalies is essential, requiring multi-category predictions to determine both the presence of an anomaly and its source. In summary, addressing multivariate time series anomaly detection in modern applications entails overcoming challenges related to data complexity, inference speed, and anomaly source tracking.

\textbf{Existing solutions.} Conventional anomaly detection methods are generally based on machine learning for processing, which can be basically subdivided into three cases: supervised learning \cite{liu2022deep}, unsupervised learning \cite{ergen2019unsupervised, shi2023robust, zhang2019delr} and weakly supervised learning \cite{zhou2021feature}, depending on the labeling of the data. However, in most real cases, the time series often lack labels, and the collected data are often streaming time series data. At the same time, labeling anomalies requires high expert cost and this does not guarantee that all anomaly types are accurately labeled, hence unsupervised learning is widely used, such as based on the statistical distribution (HBOS) \cite{goldstein2012fast}, based on distance (KNN) \cite{ramaswamy2000efficient}, based on density estimation (LOF, LFCOF) \cite{breunig2000lof, amer2012nearest}, based on clustering (EM, OC-SVM, SVDD) \cite{pan2010ganesha, scholkopf2001estimating, tax2004support}. Although there are many unsupervised methods available, they cannot handle the intrinsic features of time series data, and these classical methods do not take into account the temporal correlation, making it difficult to capture inductive information effectively in real-life scenarios. In addition, the dependencies between high-dimensional variables are mostly ignored when dealing with multi-variate time series, which decreases the precision of prediction. Today, a large number of DL-based anomaly detection methods are available, and they perform significantly better than traditional methods in solving challenging detection problems in a variety of real-world applications. Therefore, a major challenge in multivariate time series prediction is how to capture the dynamic correlation between multiple variables. In this case, the model is more adaptable in realistic scenarios. Deep learning methods use neural networks with more hidden layers and are therefore able to capture complex latent features and temporal correlations. For example, recurrent neural networks (RNN) \cite{rumelhart1986learning} have pioneered sequence modeling. However, for long sequence modeling that may lead to gradient disappearance, traditional RNN have difficulty capturing remote dependencies. As its variants, long short-term memory (LSTM) \cite{hochreiter1997long} and gated recurrent unit (GRU) \cite{cho2014learning} overcome their limitations. The crossover of the attention mechanism \cite{bahdanau2014neural} can also help RNN to model temporal patterns, and by selecting the partial sequence that focuses on the input, the dependencies of the inputs and outputs can be modeled. Time series prediction models with attention mechanisms based on LSTM or GRU have been proposed and have shown good performance in mining long-term dependencies and dealing with nonlinear dynamics \cite{shih2019temporal}. However, this structure may perform less well for dynamic periodic or aperiodic modes commonly encountered in complex environments, and recursive models such as LSTM require a large amount of computation are relatively slow in speed, and cannot continuously and effectively simulate long-term trends \cite{audibert2020usad, deng2021graph}. This is because the recursive model requires information about all previous units to be passed first when performing the next inference step.

\textbf{New approaches.} Recent developments in the Transformer model \cite{vaswani2017attention} allow for a single-step inference using a complete input sequence encoded by position, which can be detected faster than recursive methods using the Transformer by reasoning in parallel on the GPU. Moreover, the benefit of the Transformer's encoding of large sequences enables accuracy and inference time to be almost independent of sequence length. Therefore, we use the Transformer to increase the temporal contextual information without significantly increasing the computational overhead. For the above studies, many Transformer-based anomaly detection models \cite{ma2023btad, xu2021anomaly, tuli2022tranad} are proposed today with good results.

\textbf{Our model.} Although the Deep anomaly detection constructed based on Transformer has the above advantages, it has limitations in capturing long-term dependencies and the complexity of each layer of Self-Attention is a $O\left( {{n}^{2}}\cdot d \right)$, so the computational complexity of the model for long sequences is larger and the training is slower. A simple transformer-based codec may not be able to recognize little anomalies when the proportion of data anomalies is too small. Therefore, we propose a Transformer and Time Series Convolution Based Anomaly Detection Model (DTAAD). Convolutional networks can outperform RNN in many sequence modeling tasks \cite{bai2018empirical}, while avoiding common pitfalls of recursive models, such as the inability to model quickly or gradient explosion/disappearance. In addition, using a convolutional network instead of a recursive network can improve performance because it allows parallel computation of the output. TCN represents causal, dilated convolutional networks in the time domain, which consists of 1D convolutional layers with the same input and output length \cite{bai2018empirical}. TCN has many advantages, such as stable gradients, parallelism, flexible receptive fields, and lower memory, \textit{etc}. Next, we briefly describe the model architecture designed with TCN for the main components and the reasons for doing so: 1) We first send the input time series into two parallel TCN components, which are called global TCN and local TCN, to simulate the complex mode of both the global and local time patterns simultaneously. Next, the time series learned from each temporal convolutional component are sent to a separate encoder with the aim of studying the dependencies between the different series; 2) After all, TCN is a variant of the CNN, although the perceptual field can be extended using dilated convolution, but still subject to limitations, so we treat the overall model as an AR and adding AE structure combined, using the Transformer's feature of being able to capture relevant information of any length for an integrated design. Using causal convolution and dilated convolution as local TCN and global TCN, introducing scaling methods and feedback mechanisms to improve the prediction accuracy and expand the association discrepancy; 3) The perceptual field of dilation convolution is set for the global TCN, and the minimum number of convolution layers is set to ensure the global perceptual field according to the input sequence length, convolution kernel size, and dilation coefficient. Our experiments are to show that this can assign an appropriate training loss ratio for a Dual Time Series Convolution to local and global anomaly detection of the input sliding window sequence, a construction that amplifies the reconstruction error. We construct the Dual TCN-Attention Network (DTA) lightweight model with only a single-layer Transformer encoder for baseline experiments, which greatly reduces the computational complexity of the Transformer. Our extensive experiments on publicly available datasets validate that DTAAD exceeds the majority of currently advanced baseline methods in both detection and diagnostic performance. In particular, DTAAD improved F1 scores by up to $8.38\%$ and reduced training time by up to $99\%$ compared to baseline.

The main contributions of this paper are summarized as follows.

\begin{itemize}
	\item According to our knowledge, TCN is rarely introduced in deep anomaly detection designs for multivariate time and is usually used only for pure prediction. We introduce it into anomaly detection and combine it with meta-learning (MAML) \cite{finn2017model} to help maintain the best anomaly detection performance with limited data.
\end{itemize}

\begin{itemize}
	\item Here, we propose an anomaly detection framework based on local and global attention based on the architecture of AR and AE combined with the introduction of a feedback mechanism to reflect anomaly differences. 
\end{itemize}

\begin{itemize}
	\item The dilation convolution in the design of DTAAD is only set to the minimum convolution layers ensuring the global perception field. In addition, only a single-layer transformer encoder is used. This model belongs to the ultra-lightweight model, and the training time is reduced by $99\%$ compared with other baseline methods.
\end{itemize}

\begin{itemize}
	\item DTAAD obtained the most advanced anomaly detection results in benchmark tests for most real applications.
	%We employ a number of datasets to evaluate the performance of the proposed anomaly detection method. Extensive experiments are also conducted to discuss and analyze the potential benefits of the proposed method.
\end{itemize}

%demo file is intended to serve as a ``starter file''
%for IEEE journal papers produced under \LaTeX\ using	
%IEEEtran.cls version 1.8b and later.
% You must have at least 2 lines in the paragraph with the drop letter
% (should never be an issue)
%I wish you the best of success.

%\hfill mds
 
%\hfill August 26, 2015

\section{Related Works}
\label{sec:related_work}

% needed in second column of first page if using \IEEEpubid
%\IEEEpubidadjcol
\subsection{Deep Unsupervised \& Weakly Supervised Anomaly Detection}

In recent years, deep unsupervised learning~\cite{ma2023btad, wang2022variational} and deep weakly supervised learning~\cite{YangLiu2022CollaborativeNL} anomaly detection methods have been widely studied for their ability to obtain complex internal relationships directly from unlabeled data or a small part of finitely labeled data through network feature extraction learning. Most studies are based on the following three main categories: i) Deep distance-based Measure. Depth distance-based anomaly detection seeks to study characteristic representations dedicated to specific types of distance anomalies. One drawback of traditional distance-based methods is that they cannot be effective in multidimensional data, while depth-distance-based anomaly detection techniques can overcome this limitation well by first projecting the data into a low-dimensional space before distance measurement. In~\cite{pang2018learning}, a ranking model-based framework RAMODO was proposed to make a specific low-dimensional representation of ultra-high-dimensional data based on random distances. Representative Neighbors~\cite{liu2021anomaly}, which projects high-dimensional data to a low-dimensional space through sparse operations and learns to explore representative neighborhood features through self-representation. However, distance-based methods tend to be more computationally intensive, and secondly, the detection capability is limited by inherent drawbacks. ii) Deep clustering-based Measure. Anomaly detection based on deep clustering focuses on making the anomalies significantly deviate from the clustering clusters in the studied representation space, which is more effective than the traditional clustering methods. In~\cite{ruff2018deep}, Ruff et al. proposed Deep SVDD, a deep anomaly detection method, to aggregate ordinary data into a tight hypersphere. A dual self-encoder network was designed in~\cite{yang2019deep}, which robustly represents the latent features and reconfiguration constraints on noise, embedding the inputs into a latent space for clustering. However, the performance of this method for anomaly detection relies strongly on the clustering effect, and the clustering process may be misaligned due to anomalous disturbances such as noise in the training data, which can lead to a more inefficient representation. iii) Deep one-class Classification-based Measure. In this framework, the algorithm learns the single-classification discriminant boundary around the normal instances. This technique usually suffers from suboptimal performance when the dimension increases in traditional single-classification methods. A one-class neural network (OC-NN) model was proposed in~\cite{chalapathy2018anomaly} for efficient anomaly detection on complex data. OC-NN combines the rich data representation capabilities of deep network extraction to create tightly enveloped single-class targets around normal data.~\cite{shen2020timeseries} fuses the multiscale temporal features of the intermediate layers through a hierarchical clustering mechanism, which captures temporal dynamics at multiple scales using an expansive recurrent neural network with skip connections. The disadvantage of single classification is that its model may be ineffective in datasets with high-dimensional distributions within normal classes, and the detection performance depends on the anomaly measure of single classification.

\subsection{Transformers for Deep Anomaly Detection}

Many anomaly detection models~\cite{xu2021anomaly, tuli2022tranad} based on Transformer
~\cite{vaswani2017attention} have been proposed today with good results. Its ability in point-by-point representation and unified modeling of pairwise associations is shown to be tremendous, and we find that the distribution of self-attention weights at each time point can reflect rich associations with the whole sequence. In~\cite{xu2021anomaly}, a new converter of anomalous attention mechanism is technically proposed to compute association differences, and a very large and very small strategy is designed to amplify the positive anomaly discriminability of association differences.~\cite{tuli2022tranad} proposed TranAD, which uses an attention-based serial encoder to capture trends in the temporal data and make fast inferences. We use it as one of the baseline experiments since outperforms most current baselines.

In our baseline experiments, our model DTAAD is compared with the most popular and advanced TranAD~\cite{tuli2022tranad}, LSTM-NDT~\cite{hundman2018detecting}, DAGMM~\cite{zong2018deep}, OmniAnomaly~\cite{su2019robust}, MSCRED~\cite{zhang2019deep}, MAD-GAN~\cite{li2019madgan}, USAD~\cite{audibert2020usad}, MTAD-GAT~\cite{zhao2020multivariate}, CAE-M~\cite{zhang2021unsupervised}, and GDN~\cite{deng2021graph}. These approaches offer unique advantages in abnormality detection and diagnosis, but each also has disadvantages in terms of performance on various time series datasets. In summary, the main disadvantages include 1) Limited view of the local context window, which makes it unable to capture global information and incapable of detecting long-term anomalies; 2) High complexity of the model, requiring significant computational resources and training time, particularly for graph neural networks and recurrent networks; 3) Noise-sensitive, the method is prone to generate many false positives when training data contains high noise, limiting its capability for complex behavior learning. In short, it is unable to balance performance and the accuracy of global prediction in high-dimensional data. We developed the DTAAD model with a strong generalization ability for learning by using different public training datasets as a way to easier detection of anomalies in the test data. In this task, we optimize the performance bottleneck of anomaly detection with a significant reduction in training time.

\section{Background}
\label{sec:method}
\begin{table*}[!t]
	\centering %\renewcommand*{\arraystretch}{0.92}
	\caption{Glossary of terms used in the paper and their explanations}
	\resizebox{\textwidth}{!}{
	\begin{tabular}{|c|c|l|}
		\hline
		\textbf{descriptions}                                                        & \textbf{Symbol}                                                                                                  & \textbf{Meaning}                                                                                                                                                                        \\ \hline
		\begin{tabular}[c]{@{}c@{}}a set of multivariate \\ time series\end{tabular} & $\mathcal{T}=\left\{ x_{1}^{\left( i \right)},x_{2}^{\left( i \right)}\ldots ,x_{T}^{\left( i \right)} \right\}$ & It is a set of fully observed time series with a time stamp $T$.                                                                                                                        \\ \hline
		\begin{tabular}[c]{@{}c@{}}arbitrary \\ observations\end{tabular}            & ${{z}_{i,t}}$                                                                                                    & The ${{z}_{i,t}}\in {{\mathbb{R}}^{m}}$ denotes the value of $t$ time series $i$.                                                                                                       \\ \hline
		\begin{tabular}[c]{@{}c@{}}observed \\ time-dependence\end{tabular}          & $p\left( {{z}_{1:t}} \right)$                                                                                    & \begin{tabular}[c]{@{}l@{}}Observed values are not independent but exhibit some kind of structure, such as a \\ 1st-order Markov chain.\end{tabular}                                    \\ \hline
		\begin{tabular}[c]{@{}c@{}}non-linear mapping \\ probability\end{tabular}    & $p\left( \left. {{z}_{i,t}} \right|{{z}_{i,t-1}},{{x}_{i,1:t}}\text{ } \right)$                                  & Non-linear relationship of observations with potential space and inputs.                                                                                                                \\ \hline
		white noise                                                                  & ${{\epsilon }_{t}}$                                                                                              & \begin{tabular}[c]{@{}l@{}}Noise is generated according to a certain signal-to-noise ratio with the variation of \\ input time.\end{tabular}                                             \\ \hline
		anomaly score                                                                & ${{s}_{i}}$                                                                                                      & \begin{tabular}[c]{@{}l@{}}The resulting predicted value is reconstructed with the input value according to the\\ input window to produce the error as an exception score.\end{tabular} \\ \hline
		\begin{tabular}[c]{@{}c@{}}encoding network \\ parameters\end{tabular}       & ${{\phi }_{e}}\left( \hat{z};{{\Theta }_{e}} \right)$                                                            & \begin{tabular}[c]{@{}l@{}}where ${{\phi }_{e}}$ is the encoding network with the parameters combined with the input $\hat{z}$ \\ of TCN.\end{tabular}                                  \\ \hline
		\begin{tabular}[c]{@{}c@{}}decoding network \\ parameters\end{tabular}       & ${{\phi }_{d}}\left( \text{h};{{\Theta }_{d}} \right)$                                                           & \begin{tabular}[c]{@{}l@{}}where ${{\phi }_{d}}$ is the decoding network with the parameters combined with the hidden \\ state $\text{h}$ of encoder.\end{tabular}                    \\ \hline
		loss                                                                         & ${{\left\| \mathcal{O}-{{\mathcal{X}}_{t}} \right\|}_{2}}$                                                       & Calculate the loss with L2-norm.                                                                                                                                                        \\ \hline
	\end{tabular}}
	\label{tab:glossary}
\end{table*}    

\subsection{Problem Definition}
We now discuss the input sequence, which is a group of fully observed time series with a time stamp $T$
\[\mathcal{T}=\left\{ x_{1},x_{2}\ldots ,x_{T} \right\}\]
Because a given input time series is obtained after it has been fully observed, statisticians call the constant dynamics stationary. Thus, the estimates of the entire series obey the following conditional probability distribution:
\begin{equation}
	p(x)=\underset{t=1}{\mathop{\overset{T}{\mathop{\prod }}\,}}\,p\left( \left. x_{t} \right|x_{1},\ldots ,x_{t-1} \right)
\end{equation}
Given an input multivariate time series \[\mathcal{T}=\sum\nolimits_{i=1}^{N}{\sum\nolimits_{l=1}^{m}{\left\{z_{i,1:{{t}_{0}}}^{l} \right\}}}\]
where $z_{i,1:{{t}_{0}}}^{l}\triangleq \left[ z_{i,1}^{l},z_{i,2}^{l},...,z_{i,1:{{t}_{0}}}^{l} \right]$ and $z_{i,1:{{t}_{0}}}^{l}\in {{\mathbb{R}}^{m}}$ denotes the value of time series $i$ in the $l$-th dimension at time $t$ where each datapoint $x_{t}\in {{\mathbb{R}}^{m}},\forall t$. Here, $N$ represents the number of samples in the dataset and $m$ refers to the number of time dimensions. Our goal is to make the model conform to the following conditional probability  distribution:
\begin{equation}
	\resizebox{0.91\hsize}{!}{$\begin{aligned}
			p\left( \left. \text{z}_{i,{{t}_{0}}+1:{{t}_{0}}+\tau }^{l} \right|\text{z}_{i,1:{{t}_{0}}}^{l},{{\mathbf{x}}^{l}_{i,1:{{t}_{0}}+\tau }};\text{ }\Theta \text{ } \right)=\underset{t={{t}_{0}}+1}{\mathop{\overset{{{t}_{0}}+\tau }{\mathop{\prod }}\,}}\,p\left( \left. {{\mathbf{z}}^{l}_{i,t}} \right|{{\mathbf{z}}^{l}_{i,1:t-1}},{{\mathbf{x}}^{l}_{i,1:t}};\text{ }\Theta \text{ }\right)
		\end{aligned}$}
\end{equation}
We will predict the time series for the next $\tau$ time steps, because single-step prediction is convenient for experiments, our subsequent baseline experiments let the model learn the parameters $p\left( {\left. {z_{i,t}^l} \right|z_{i,1:t - 1}^l,x_{i,1:t}^l;{\rm{ }}\Theta {\rm{ }}} \right)$ based on single-step prediction. Since unsupervised learning does not have well-labeled data, our main goal is to achieve the following two effects:

\begin{itemize}
	\item \textbf{Anomaly detection}, \textit{i.e.}, For an input training time series $\mathcal{T}$, our aim is to use the sliding window for subsequent $h$-step predictions, where $h\ge1$. We predict that ${{x}_{T+h}}$ based on the known $\mathcal{T}$, we need to predict the value of $\mathcal{Y}=\left\{ {{y}_{T}},\ldots,{{y}_{T+h}} \right\}$, ${{y}_{t}}\in \left\{ 0,1 \right\}$ for detecting anomaly events at $h$ time steps after $T$. Similarly, we predict the next sequence of ${{x}_{T+h+k}}$ based on $\left\{ {{x}_{1+k}}^{(i)},{{x}^{(i)}}_{2+k},\ldots,{{x}_{T+k}}^{(i)} \right\}$, $k\in {{\mathbb{R}}^{+}}$, The sliding window size is generally set to a fixed value.
\end{itemize}
\begin{itemize}
	\item \textbf{Anomaly diagnosis}, \textit{i.e.}, Global qualitative analysis of each dimension is performed based on the anomaly detection results of all input timings, and the time period of each anomaly is accurately identified to analyze the severity of the anomaly. we will forecast $\mathcal{Y}=\left\{ y_{1}^{\left( i \right)},y_{2}^{\left( i \right)},\ldots,y_{T+h}^{\left( i \right)} \right\}$, where ${{y}_{t}}\in {{\left\{ 0,1 \right\}}^{m}}$ indicates which patterns of the data point are anomalous at the $\tau $ time tamp.
\end{itemize}

\subsection{Data Preprocessing}

In the real world, due to various natural or extreme events and other different reasons, the collected time series data are often disturbed by some noise in addition to system failures, \textit{etc}. To enable our model training to be more effective, we standardize the training and test data and add white noise with a signal-to-noise ratio of 50 dB as data enhancement. Experimental testing demonstrated that the addition of appropriate noise did indeed enhance the generalization ability and robustness of the model against noise interference, and finally slice into sliding time windows

\begin{equation}
	{{\hat{x}}_{t}}\leftarrow \frac{{{x}_{t}}-\min \left( \mathcal{T} \right)}{\max \left( \mathcal{T} \right)-\min \left( \mathcal{T} \right)+\epsilon }
\end{equation}
\begin{equation}
	SNR=10{{\log }_{10}}\frac{\sum\nolimits_{1}^{N}{{{\left( {{{\hat{x}}}_{t}} \right)}^{2}}}}{\sum\nolimits_{1}^{N}{{{\left( {{\epsilon }_{t}} \right)}^{2}}}}
\end{equation}
\begin{equation}
	{{x}_{t}}={{\hat{x}}_{t}}+\frac{{{\epsilon }_{t}}}{100}
\end{equation}
where ${{\hat{x}}_{t}}$ and ${{\epsilon }_{t}}$ denote the signal and noise, $N$ represents the number of samples in the dataset, and the signal-to-noise ratio SNR is in dB. where $\max \left( \mathcal{T} \right)$ and  $\min \left( \mathcal{T} \right)$ are the minimum and maximum vectors in the training time-series $\mathcal{T}$. $\epsilon$ is a very small constant added to prevent divide-by-zero overflow, and the entire input data has a range of $\left( 0,{{1}^{+}} \right)$. We consider a fixed local contextual window of length $K$ as
${{\mathcal{W}}_{t}}=\left\{ {{x}_{t-K+1}},\ldots,{{x}_{t}} \right\}$. We use copy fill for $t<K$ and transform the input time series $\mathcal{T}$ into the sliding windows $\mathcal{W}=\left\{ {{\mathcal{W}}_{1}},\ldots,{{\mathcal{W}}_{t}} \right\}$. We use $\mathcal{W}$ as the model training, and $\hat{\mathcal{W}}$ as the test sequences to replace $\mathcal{T}$. Our aim is to generate reconstruction errors from the predicted values to the original values based on the input window, which are then used as anomaly scores ${{s}_{i}}$. Here, we select the anomaly threshold automatically using the Peak Over Threshold (POT)~\cite{siffer2017anomaly} approach and then combine all abnormal labels ${{y}_{i}}\left( {{s}_{i}}\ge \text{POT}\left( {{s}_{i}} \right) \right)$ into a total abnormal distribution:
\begin{equation}
	D=\underset{i}{\mathop{\vee }}\,{{y}_{i}}
\end{equation}

\section{Methodology}
\begin{figure*}
	\centering %\setlength{\belowcaptionskip}{-10pt}
	\includegraphics[width=\linewidth]{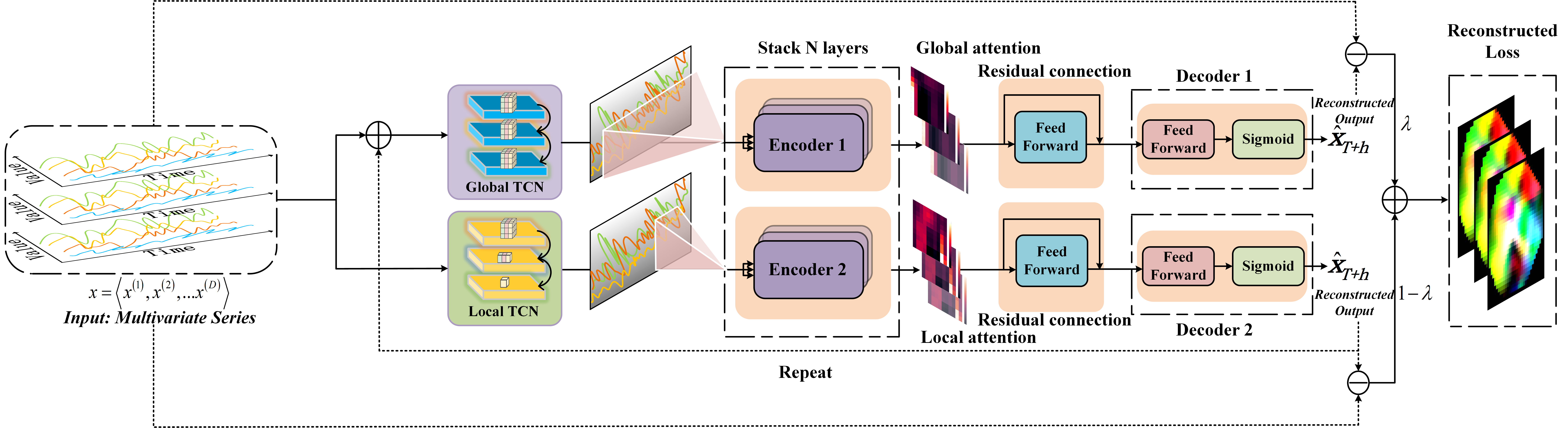}
	\caption{Dual Tcn-Attention Network (DTANet). This model mainly consists of two parts, \textit{i.e.}, a Tcn-based approximate autoregression layer and a Transformer-based encoder-decoder layer. Among them, TCN is generally a fixed layer in order to satisfy the receptive field, and the encoding layer can be integrated multiple times. Different attentions are sent to the decoder through the residual connection and the final prediction of the local attentions is sent back to the global TCN together with the original input overlay by a copy operation, finally, the two losses are reconstructed according to a certain proportion $\lambda$.}
	\label{fig:model}
\end{figure*}

\subsection{Model Framework}
% Models like LSTM and GRU are often slow to train, and in an effort to eschew using recurrence based models, we use a transformer based data reconstruction approach.
\emph{\textbf{Overall Architecture.}} Figure~\ref{fig:model} illustrates our overall model DTAAD. Our overall model can be approximated as an autoregressive model (AR). The first half of the model consists of dual TCN, and the codec layer model (AE) is built based on the Transformer in the back. DTAAD utilizes two temporal convolutional structures, causal convolution, and dilated convolution as local TCN and global TCN, which will generate the same number of temporal prediction outputs as the input, and then into a Transformer encoding layer for capturing the dependencies between multiple series. Then, the global attention and local attention output from the coding layer are flowed into a specific decoding layer according to the residual connection~\cite{he2016deep}, and the final prediction results of the local attention are feedback to the global TCN and the original input overlay by the replication operation, and finally, the reconstruction errors obtained based on the two sets of prediction values are combined in a certain proportion to obtain the training loss. The following paragraphs describe the details of the building blocks one by one.

\begin{figure}
	\centering %\setlength{\belowcaptionskip}{-10pt}
	\includegraphics[width=0.7\linewidth]{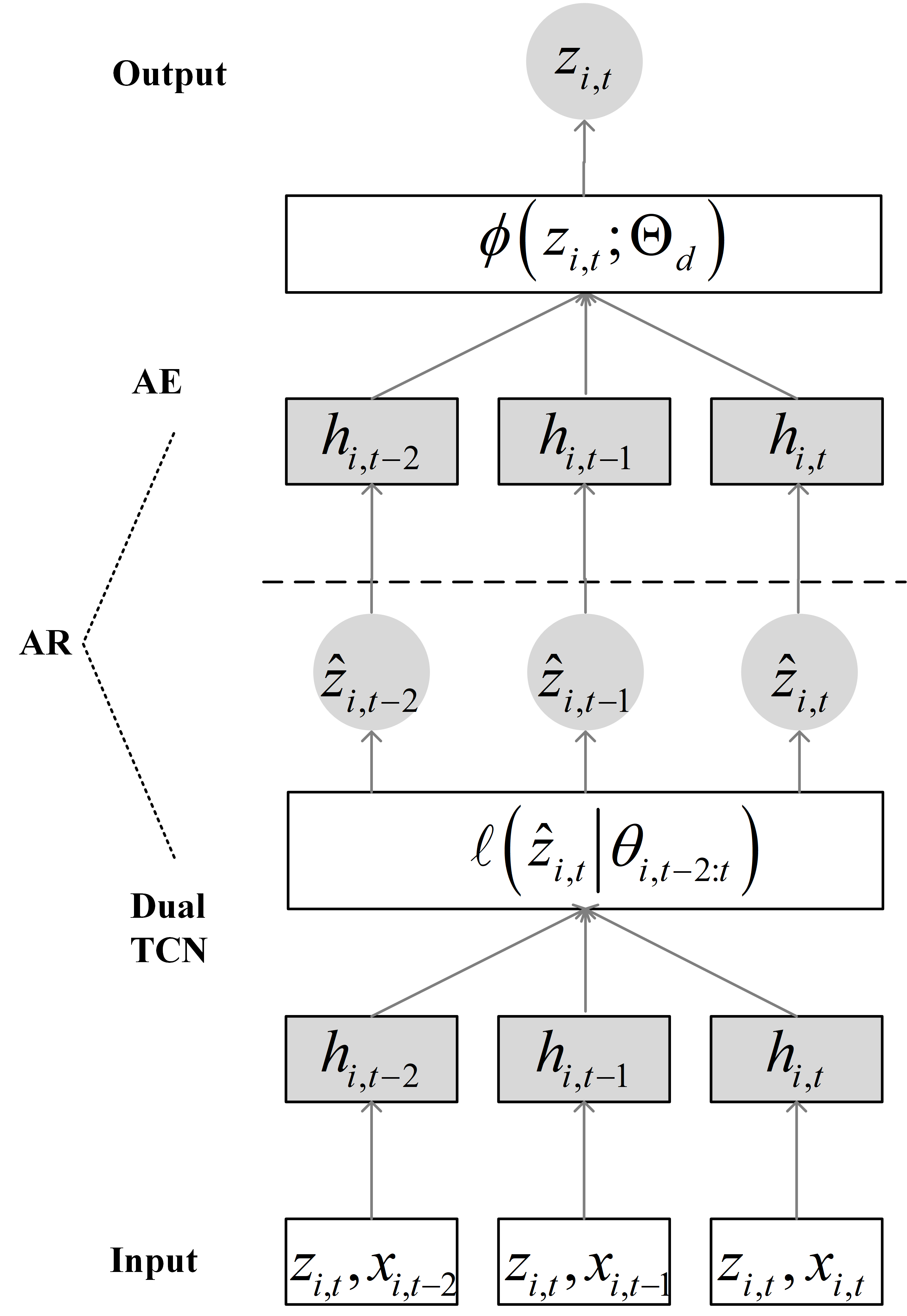}
	\caption{ Overall computational flow. The likelihood $\ell \left( \left. {{{\hat{z}}}_{i,t}} \right|{{\theta }_{i,t-2:t}} \right)$ is a distribution whose parameters ${{\theta }_{i,t-2:t}}$ are given by the Dual TCN hidden state ${{\text{h}}_{i,t-2:t}}$. The $\phi \left( {{z}_{i,t}};{{\Theta }_{d}} \right)$ is the decoder network that aggregates the encoder inputs to the final output ${{z}_{i,t}}$.}
	\label{fig:computational flow}
\end{figure}

Our network, concluded in Fig~\ref{fig:model}, is an approximate autoregressive architecture. We suppose that our AR network distribution ${{Q}_{\Theta }}\left( \left. {{z}_{i,{{t}_{0}}:T}} \right|{{z}_{i,1:{{t}_{0}}-1}},{{x}_{i,1:T}} \right)$ consists of a product of likelihood factors:
\begin{equation}
	\label{eq:likelihood}
	\resizebox{0.91\hsize}{!}{$\begin{aligned}
			{{Q}_{\Theta }}\left( \left. {{z}_{i,{{t}_{0}}:T}} \right|{{z}_{i,1:{{t}_{0}}-1}},{{x}_{i,1:T}} \right)=\underset{t={{t}_{0}}}{\mathop{\overset{T}{\mathop{\prod }}\,}}\,{{Q}_{\Theta }}\left( \left. {{z}_{i,t}} \right|{{z}_{i,1:t-1}},{{x}_{i,1:T}} \right)=\underset{t={{t}_{0}}}{\mathop{\overset{T}{\mathop{\prod }}\,}}\,\ell \left( \left. {{z}_{i,t}} \right|\theta \left( {{\text{h}}_{i,1:t}},\Theta  \right) \right)
		\end{aligned}$}
\end{equation}
parametrized by the hidden state ${{\text{h}}_{i,t}}$ of the AR model:
\begin{equation}
	\label{eq:AR}
	{{\text{h}}_{i,t}}=h\left( {{x}_{i,t}},\Theta  \right)
\end{equation}	
In~\eqref{eq:likelihood}, the likelihood $\ell \left( \left. {{z}_{i,t}} \right|\theta \left( {{\text{h}}_{i,1:t}},\Theta  \right) \right)$ is a fixed distribution whose parameters are given by a function $\theta \left( {{\text{h}}_{i,1:t}},\Theta  \right)$ of the AR model hidden state ${{\text{h}}_{i,t}}$, $\Theta$ is the AR model parameter, which strictly consumes all previous hidden states ${{\text{h}}_{i,1:t}}$. $h$ in~\eqref{eq:AR} is the functional function implemented in the AR section, as detailed in the training computational flow diagram referable in Fig.\ref{fig:computational flow}. The basic formula for the codec part is as follows:
\begin{equation}
	\text{h}={{\phi }_{e}}\left( \hat{z};{{\Theta }_{e}} \right),z={{\phi }_{d}}\left( \text{h};{{\Theta }_{d}} \right)
\end{equation}
\begin{equation}
	\resizebox{0.89\hsize}{!}{$\begin{aligned}
	\left\{ \text{ }\!\!\Theta\!\!\text{ }_{e}^{*},\text{ }\!\!\Theta\!\!\text{ }_{d}^{*} \right\}=\underset{{{\text{ }\!\!\Theta\!\!\text{ }}_{e}},{{\text{ }\!\!\Theta\!\!\text{ }}_{d}}}{\mathop{\arg \min }}\,\frac{{{\left( \sum\limits_{{{x}_{t}}\in \mathcal{X}}{{{x}_{t}}-{{\phi }_{d}}\left( {{\phi }_{e}}\left( \hat{z};{{\Theta }_{e}} \right);{{\text{ }\!\!\Theta\!\!\text{ }}_{d}} \right)} \right)}^{2}}}{n}
	\end{aligned}$}
\end{equation}
\begin{equation}
	{{s}_{t}}={{\left( {{x}_{t}}-{{\phi }_{d}}\left( {{\phi }_{e}}\left( x;\Theta _{e}^{*} \right);\Theta _{d}^{*} \right) \right)}^{2}}
\end{equation}
where ${{\phi }_{e}}$ is the encoding network with the parameters ${{\Theta }_{e}}$ and ${{\phi }_{d}}$ is the decoding network with the parameters ${{\Theta }_{d}}$. $\hat{z}$ is the output of the AR part.$n$ is the number of total datasets. ${{s}_{t}}$ is a reconstruction error-based anomaly score of ${{x}_{t}}$.

\emph{\textbf{Dual TCN.}} The overall part consists of two parts, Local Tcn and Global Tcn. TCN is dependent on two main criteria: the output length of the network is the equal to the input, and there is no leakage of information from the past to the future. Input time series $\mathcal{X}\in {{\mathbb{R}}^{N\times d}}$. Excluding the fill-and-crop operation in TCN, its single-layer basic computational flow is as follows:
\begin{equation}
	\mathcal{Z}=Weight\text{-}Norm\left( Conv1D\left( \mathcal{X} \right) \right)
\end{equation}
\begin{equation}
	\mathcal{F}=Dropout\left( Leaky\operatorname{Re}lu\left( \mathcal{Z} \right) \right)
\end{equation}
where $\mathcal{X}=\left\{ x_{1}^{\left( i \right)},x_{2}^{\left( i \right)}\ldots ,x_{T}^{\left( i \right)} \right\}$ with ${{d}_{\text{model}}}$ channels. The $\mathcal{Z}$ is output after 1D convolutional network and Weight normalization~\cite{salimans2016weight}. The $\mathcal{F}$ is the final output of passing $\mathcal{Z}$ through LeakyRelu
~\cite{maas2013rectifier} and Dropout~\cite{srivastava2014dropout} where $Leaky\operatorname{Re}lu=\max \left( 0,\mathcal{Z} \right)+leak\cdot \min \left( 0,\mathcal{Z} \right)$, the $leak$ is a very small constant. Now we introduce these two Tcn structures one by one (see below). Local Tcn: Figure~\ref{fig:local} demonstrates its basic design structure.

Due to the excellent ability of convolutional structures in capturing features and parallel computation, we apply causal convolution~\cite{oord2016wavenet} to our local temporal convolution section. For causality, given an input sequence $\mathcal{T}=\left\{ {{x}_{1}},{{x}_{2}}\ldots ,{{x}_{T}} \right\}$, the $i$-th element ${{x}_{t}}$ of its output sequence may be based on only a small segment of the input sequence. Obviously, the elements in the output sequence can only depend on the elements whose index precedes  it in the input sequence. As mentioned before, to guarantee that an output tensor has the same length as the input tensor, we need to perform zero padding. The zero padding length is (Kernel -1) to maintain all layers at the same length as the input. As we know from the figure, for each element in the output sequence, its latest dependency in the input sequence has the same index as itself. So in causal convolution, the input ${{x}_{t}}$ at moment $t$ can only reach the current layer output ${{\mathcal{X}}^{t,l}}$ at moment $t$ , and the output is convolved with the earlier elements in the previous layer 
\begin{equation}
	{{\mathcal{X}}^{t,l}}=f\left( {{W}^{l}}*{{\mathcal{X}}^{t-k+1:t,l-1}}+{{b}^{l}} \right)
\end{equation}
where $*$ denotes the convolutional operation, $f\left( \cdot  \right)$ is the activation function, ${{W}^{l}}$ denotes convolutional kernels of size $k$, ${{b}^{l}}$ is a bias term. Since models with causal convolution do not have circular transfer, their training speed is superior to that of RNN. However, causal convolutions also have shortcomings, they usually need many layers to increase the receptive field, are computationally expensive for long sequences, and are prone to gradient vanishing. Therefore, we only use it as a local detection, generally for a fixed number of shallow layers. In the following presentation, we use dilation convolution to increase the order of magnitude of the receiver field as global detection without adding much computational cost. Global Tcn: Figure~\ref{fig:Global} demonstrates its basic design structure.

\begin{figure}
	\centering %\setlength{\belowcaptionskip}{-10pt}
	\includegraphics[width=\linewidth]{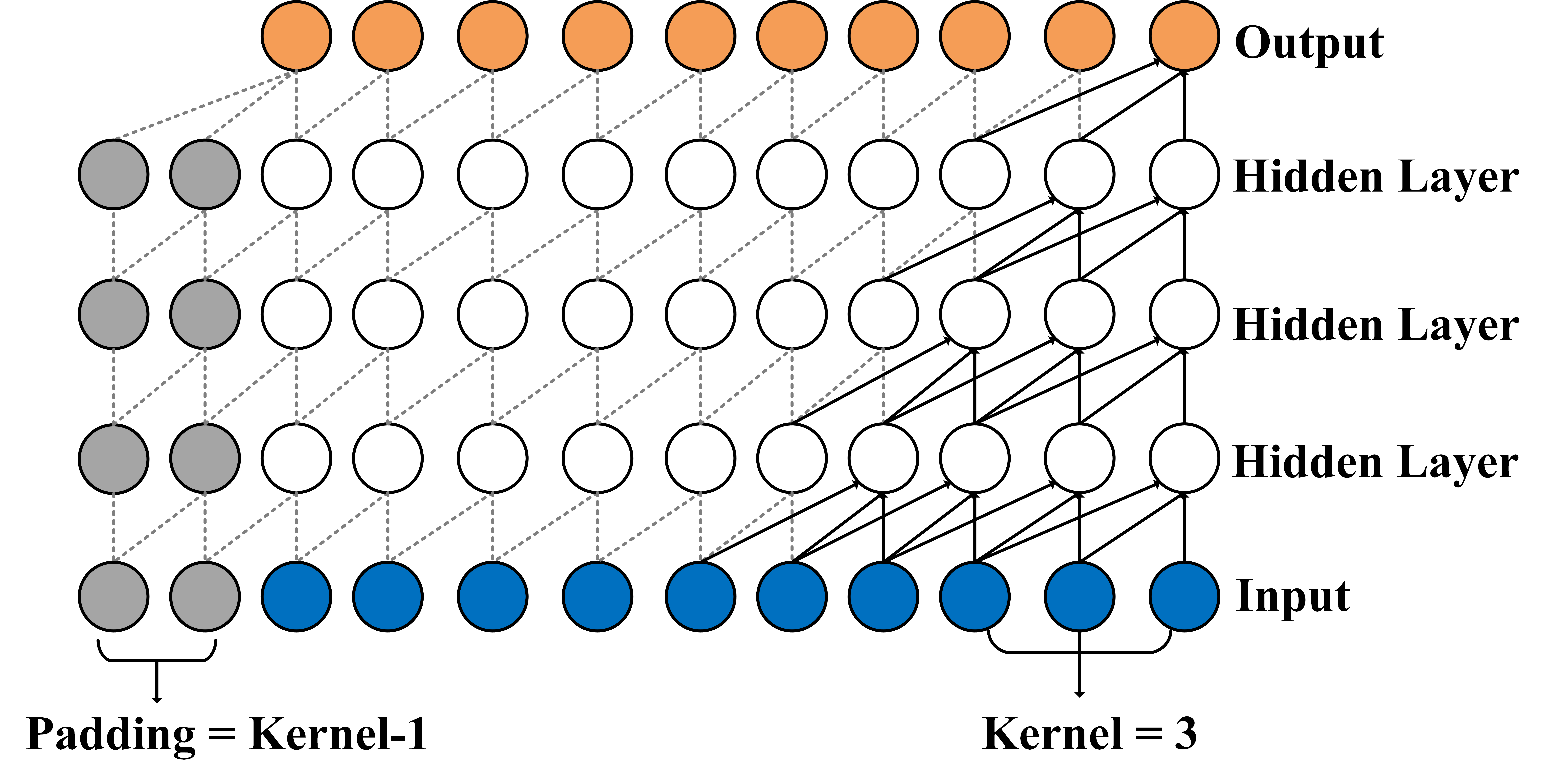}
	\caption{ Local TCN (Causal Convolutions). This is the structure consisting of three sets of hidden layers, The convolution kernel of size $k=3$, Each layer pads $k-1$ inputs from the leftmost.}
	\label{fig:local}
\end{figure}
\begin{figure}
	\centering %\setlength{\belowcaptionskip}{-10pt}
	\includegraphics[width=\linewidth]{Global.png}
	\caption{ Global TCN (Dilated Convolutions). This is the structure consisting of three sets of hidden layers, The convolution kernel of size $k=3$, Each layer pads ${{b}^{n-1}}\cdot \left( k-1 \right)$ inputs from the leftmost.}
	\label{fig:Global}
\end{figure}

One desirability of the predictive model is that the value of a particular observation in the output depends on all previous inputs, \textit{i.e}., all entries whose index is less than or equal to itself. This is possible when the size of the convolution field is input-length, which we also call "complete history". Next, we analyze our design: using dilation convolution as global receptive field detection.

The dilation convolution is equivalent to a jump filter~\cite{oord2016wavenet}, where each layer is expanded to achieve exponential receptive field expansion. For a 1-D sequence input $\text{x}\in {{\mathbb{R}}^{n}}$ and the filter $f\text{ =}\left\{ 0,\ldots,k-1 \right\}\in \mathbb{R}$,the dilated convolution operation $F$ on the element $s$ of the sequence is defined as~\cite{bai2018empirical}
\begin{equation}
	\mathcal{F}\left( s \right)=\left( \text{x}{{*}_{d}}f \right)\left( s \right)=\sum\limits_{i=0}^{k-1}{f\left( i \right)}\cdot {{\text{x}}_{s-d\cdot i}}
\end{equation}
where $d$ is the dilation factor, $k$ is the convolutional filter size, $s-d\cdot i$ indicates the index to the past according to $d$. $d$ is equivalent to jumping a fixed step size between two adjacent layers.In general, the received field $r$ of a 1D convolutional network with $n$ layers and a kernel-size of $k$ is
\begin{equation}
	r=1+n\cdot \left( k-1 \right)
\end{equation}
We set the size of the receptive field to the input-length and solve for the number of layers $n$ such that it completely covers
\begin{equation}
	n=\left\lceil {\left( l-1 \right)}/{\left( k-1 \right)}\; \right\rceil 
\end{equation}
where $\left\lceil \cdot  \right\rceil $ is rounded up. We found that with a fixed kernel size, the number of layers required for complete history coverage is a tensor of the linear input length, which causes the network to become very deep, resulting in a model with a large number of parameters. From Figure~\ref{fig:Global}, we see that the final output value depends on the entire input coverage. In general, the current receptive field width increases by $d\cdot \left( k-1 \right)$ for each additional layer, where $d={{b}^{n-1}}$, is the expansion base $b$, and $n$ is the count of convolution layers. Thus, the width of the receptive field $w$ based on the $b$-exponential expansion, the kernel size $k$, and the number of layers $n$ for the time-dilated convolution is
\begin{equation}
	w=1+\sum\limits_{i=1}^{n}{\left( k-1 \right)\cdot {{b}^{i-1}}=1+}\left( k-1 \right)\cdot \frac{{{b}^{n}}-1}{b-1} 
\end{equation}
Further analysis shows us that the range of the acceptance field is indeed larger than the size of the input. However, the acceptance field is holed, \textit{i.e}., there are entries in the input sequence on which the output value does not depend. Given an expansion base $b$, a kernel size $k$, where $k\ge b$ and an input length $l$, in order to cover the entire input history, the following inequalities must be satisfied:
\begin{equation}
	\left( k-1 \right)\cdot \frac{{{b}^{n}}-1}{b-1}\ge l 
\end{equation}
Thus we can solve for $b$, to obtain the minimum number of layers we need to design the global TCN:
\begin{equation}
	n=\left\lceil lo{{g}_{b}}\left( \frac{\left( l-1 \right)\cdot \left( b-1 \right)}{\left( k-1 \right)}+1 \right) \right\rceil
\end{equation}
It has a computational complexity of $O\left( k\cdot l\cdot {{m}^{2}} \right)$ per layer. Moreover, we only need to stack to the minimum number of layers of the computation for the model to be able to access the information of each input. Therefore, the total cost of memory usage is only $O\left( k\cdot l\cdot n \right)$.

\begin{figure}
	\centering %\setlength{\belowcaptionskip}{-10pt}
	\includegraphics[width=\linewidth]{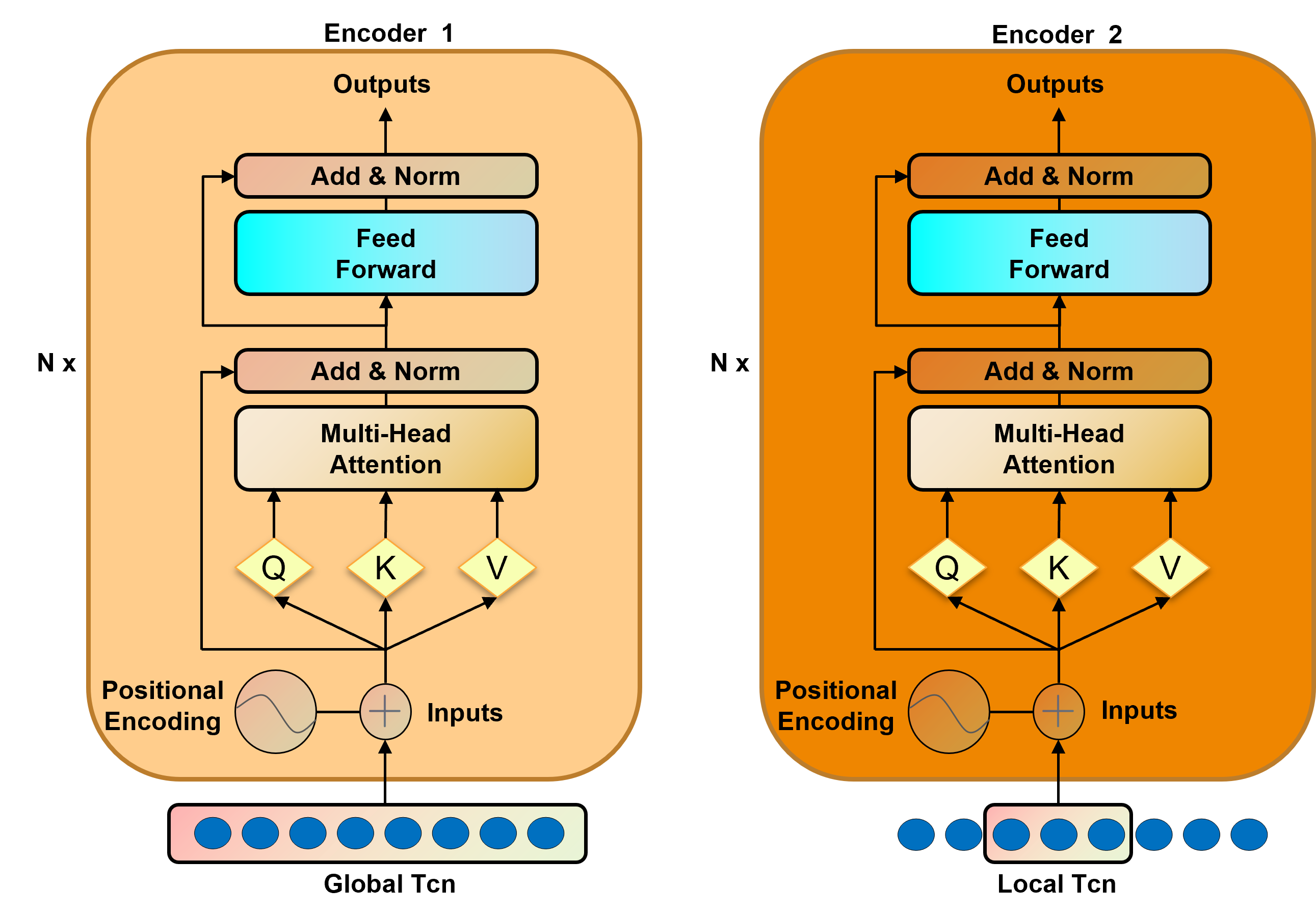}
	\caption{ Transformer (Encoder). Each base layer of the Encoder contains two sub-layers. The first sub-layer is a multi-head attention mechanism, which takes as its input $q,k,v$, respectively, from the outputs of global and local temporal convolutions, as well as the superposition of corresponding positional encoding. The second sublayer is a fully connected feedforward neural network. Residual connections and layer normalization~\cite{ba2016layer} are introduced for both sublayers.}
	\label{fig:Encoder}
\end{figure}

\emph{\textbf{Transformer Encoder}} Figure~\ref{fig:Encoder} shows the basic structure.
Transformer, as the most popular deep learning model in recent years~\cite{vaswani2017attention}, has gained significant advantages in numerous fields like natural language processing and machine vision. We will then briefly describe its architecture, which is typically designed to be stackable and integrated. However, when trying to decrease the model parameters, only its single coding layer is introduced in our baseline experiments. The core of the Transformer is the attention, Its attention-scoring function takes a dot product attention, and the input contains ${{d}_{k}}$-dimensional query and key, and ${{d}_{v}}$-dimensional value. Typically, in principle, we copy and map the inputs into queries and key-value pairs and pack them into three matrices $\mathcal{Q}$ (query), $\mathcal{K}$ (key), and $\mathcal{V}$ (value). We calculate the attention-scoring function output matrix as:
\begin{equation}
	Attention(Q, \mathcal{K}, \mathcal{V})=softmax\left(\frac{Q \mathcal{K}^T}{\sqrt{d_k}}\right) \mathcal{V}
\end{equation}
where $Q,\mathcal{K},\mathcal{V}\in {{\mathbb{R}}^{n\times {{d}_{\text{model}}}}}$. We take here $h={{d}_{\text{model}}}$, which is learned separately for different dimensions of time, and we can use the different linear projections of the $h$ sets obtained by independent learning to transform them. This $h$ group of transformed queries, keys and values is then sent in parallel to the attention aggregation, which we call multi-headed attention. It enables the model to focus on spatially diverse information at different embedding locations simultaneously. The final transformation is performed by another linear projection layer. The matrix of learnable parameters is mapped here, $\text{W}_{i}^{Q}\in {{\mathbb{R}}^{{{d}_{\text{model}}}{{d}_{k}}}},{{\text{W}}_{i}}^{\mathcal{K}}\in {{\mathbb{R}}^{{{d}_{\text{model }}}{{d}_{k}}}},\text{W}_{i}^{\mathcal{V}}\in {{\mathbb{R}}^{{{d}_{\text{model }}}{{d}_{v}}}}$.
\begin{equation}
	hea{{d}_{i}}=Attention(Q\text{W}_{i}^{Q},\mathcal{K}\text{W}_{i}^{\mathcal{K}},\mathcal{V}\text{W}_{i}^{\mathcal{V}})
\end{equation}
\begin{equation}
	\resizebox{0.89\hsize}{!}{$\begin{aligned}
	MultiHead(Q,\mathcal{K},\mathcal{V})=Concat(hea{{d}_{1}},...,hea{{d}_{h}}){{\text{W}}^{O}}
	\end{aligned}$}
\end{equation}
The second sublayer is a fully connected feedforward neural network. It consists of two linear transformations and an activation function in the middle, here we choose LeakyRelu~\cite{maas2013rectifier} as the activation function. In addition, we use a positional encoding of the sine and cosine for the input matrix~\cite{vaswani2017attention}.
\begin{equation}
	\resizebox{0.89\hsize}{!}{$\begin{aligned}
			FFN(\hat{\mathcal{Z}})=\left( max(0,\mathcal{Z}{{\text{W}}_{1}}+{{b}_{1}})+leak\cdot min\left( 0,\mathcal{Z}{{\text{W}}_{1}}+{{b}_{1}} \right) \right){{\text{W}}_{2}}+{{b}_{2}}
		\end{aligned}$}
\end{equation}
The second half of our model consists mainly of two encoders and two decoders (Fig.~\ref{fig:model}). We combine the output of the first half of the model with the positional encoding to obtain the input ${{I}_{i}}$, which is fed into two separate encoders:
\begin{align}
	\begin{split}
		& I_{i}^{1}=Layer\text{-}Norm\left( {{I}_{i}}+MultiHead({{I}_{i}},{{I}_{i}},{{I}_{i}}) \right) \\ 
		& I_{i}^{2}=Layer\text{-}Norm\left( I_{i}^{1}+FFN(I_{i}^{1}) \right)
	\end{split}
\end{align}
where $i\in \left\{ 1,2 \right\}$for the first and second encoder. The output of the encoder is then residual-connected to the feedforward layer and sent separately to the two decoders to obtain the final two predicted outputs:
\begin{align}
	\begin{split}
		& I_{i}^{3}=I_{i}^{2}+FFN\left( I_{i}^{2} \right) \\ 
		& {{\mathcal{O}}_{i}}=Sigmoid\left( FFN\left( I_{i}^{3} \right) \right) \\ 
	\end{split}
\end{align}
Sigmoid activation makes the output range ${{\mathcal{O}}_{i}}\in \left[ 0,1 \right]$ to perform the later error reconstruction with the normalized sliding window input. Figure~\ref{fig:computational flow}: The basic computational flow diagram of the overall model is shown.

\subsection{Training Procedure}
% algo, prediction outputs, minimax game, adversarial training, loss, reasoning.

\begin{algorithm}[!t]
	\begin{algorithmic}[1]
		\Require
		\Statex Training time series $\text{X}\in {{\mathbb{R}}^{T\times M}}$
		\Statex $\mathcal{W}$ for training sliding windows
		\Statex Evolutionary hyperparameter $\epsilon$
		\Statex Split time series into the dataset $\mathcal{D}={{\left\{ {{\mathcal{X}}_{d}} \right\}}_{1:D}},\mathcal{X}\in {{\mathbb{R}}^{\mathcal{W}\times M}}$
		\Statex Encoders ${{E}_{1}}$ and ${{E}_{2}}$, Decoders ${{D}_{1}}$ and ${{D}_{2}}$
		\Statex Set hyperparameters $\text{W}$
		\Statex Iteration number $N$
		\State Randomly initialize ${{\Theta }_{e}},{{\Theta }_{d}}$
		\State $n\leftarrow 0$
		\State \textbf{While} $n < N$
		\State \hspace{8pt} Randomly sample one batch from the dataset $\mathcal{D}$
		\State \hspace{8pt} \textbf{for} $t = 1 \text{ to } \mathcal{W}$ \textbf{do}
		\State \hspace{15pt} ${{\mathcal{O}}_{1}}\leftarrow \varphi _{\phi }^{{{D}_{1}}}\left( \phi \left( {{\mathcal{X}}_{:,t}};{{\Theta }_{{{E}_{1}}}} \right);\text{W} \right)$
		\State \hspace{15pt} ${{\mathcal{O}}_{2}}\leftarrow \varphi _{\phi }^{{{D}_{2}}}\left( \phi \left( {{\mathcal{X}}_{:,t}};{{\Theta }_{{{E}_{2}}}} \right);\text{W} \right)$
		\State \hspace{15pt} ${{\hat{\mathcal{O}}}_{1}}\leftarrow \varphi _{\phi }^{{{D}_{1}}}\left( \phi \left( {{\mathcal{X}}_{:,t}}+{{\mathcal{O}}_{2}};{{\Theta }_{{{E}_{1}}}} \right);\text{W} \right)$
		\State \hspace{15pt} ${{\mathcal{L}}_{1}}={{\left\| {{{\hat{\mathcal{O}}}}_{1}}-{{\mathcal{X}}_{t}} \right\|}_{2}},{{\mathcal{L}}_{2}}={{\left\| {{\mathcal{O}}_{2}}-{{\mathcal{X}}_{t}} \right\|}_{2}}$ \label{line:l1}
		\State \hspace{15pt} $\mathcal{L}\left( {{\Theta }_{e}},{{\Theta }_{d}} \right)=\lambda {{\mathcal{L}}_{1}}+\left( 1-\lambda  \right){{\mathcal{L}}_{2}}$ \label{line:l}
		\State \hspace{15pt} Compute the gradient ${{\nabla }_{\theta }}\mathcal{L}\left( {{\Theta }_{e}},{{\Theta }_{d}} \right)$ and update the 
		
		network parameter $\left\{ {{\Theta }_{e}},{{\Theta }_{d}} \right\}$
		\State \hspace{8pt} $n \gets n + 1$
		\State \hspace{8pt} Meta-Learn weights $E_1, E_2, D_1, D_2$ using a random 
		
		\hspace{-11pt} batch \label{line:meta}
		\Ensure $\phi \left( \cdot ;{{\Theta }_{e}};{{\Theta }_{d}} \right)$;
	\end{algorithmic}
	\caption{Network Training Procedure}
	\label{alg:training}
\end{algorithm}

\emph{\textbf{Objective Function.}} Our overall goal is to jointly optimize the two parts of the model. We calculate the error of the output prediction of each of the two decoders with respect to the original window input ${{x}_{t}}$, and introduce the MSE as our loss criterion:
\begin{align}
	\begin{split}
		& {{\mathcal{L}}_{1}}=\frac{\sum\nolimits_{i=1}^{n}{{{\left( {{O}_{1}}-{{x}_{i}} \right)}^{2}}}}{n} \\ 
		& {{\mathcal{L}}_{2}}=\frac{\sum\nolimits_{i=1}^{n}{{{\left( {{O}_{2}}-{{x}_{i}} \right)}^{2}}}}{n}
	\end{split}
\end{align}
Our second part combines the generated losses of the two decoders in a balanced manner in a certain hyperparameter ratio $\lambda \in \left( 0,1 \right)$ to obtain the total loss:
\begin{equation}
	\mathcal{L}\left( \psi \left( \phi \left( \cdot ;{{\Theta }_{e}};{{\Theta }_{d}} \right);\text{W} \right) \right)=\lambda {{\mathcal{L}}_{1}}+\left( 1-\lambda  \right){{\mathcal{L}}_{2}}
\end{equation}
Our ultimate goal is to minimize the total loss:
\begin{equation}
	\left\{ {{\Theta }^{*}},{{\text{W}}^{*}} \right\}=\underset{\text{ }\!\!\Theta\!\!\text{ },\text{W}}{\mathop{\arg \min }}\,\sum\limits_{x\in \mathcal{X}}{\mathcal{L}\left( \psi \left( \phi \left( \cdot ;{{\Theta }_{e}};{{\Theta }_{d}} \right);\text{W} \right) \right)}
\end{equation}
where $\phi$ represents the overall network, ${{\Theta }_{e}},{{\Theta }_{d}}$ are the total model parameters. $\text{W}$ stands for the collection of hyperparameters, and $\psi$ represents the overall learning mapping under the task of anomaly detection.

\textbf{Meta Learning} The anomaly rate of real-world collected data tends to be low, which makes the training of the model often under-fitted. We use model-agnostic meta learning (MAML)\cite{finn2017model} to do data augmentation, which helps us to better the training effect with limited data. Formally, it is the training process in which a single gradient update is recorded for the model ${{f}_{\theta }}$ with parameters $\theta$ in each random batch task ${{\mathcal{T}}_{i}}$:
\begin{equation}
	{{{\theta }'}_{i}}=\theta -\alpha {{\nabla }_{\theta }}\mathcal{L}{{}_{{{\mathcal{T}}_{i}}}}\left( {{f}_{\theta }} \right)
\end{equation}
where $\alpha$ is the learning rate and the total update of multiple gradients using multiple tasks after each training phase is completed:
\begin{equation}
	\theta =\theta -\beta {{\nabla }_{\theta }}\sum\limits_{{{\mathcal{T}}_{i}}\sim p\left( \mathcal{T} \right)}{\mathcal{L}{{}_{{{\mathcal{T}}_{i}}}}\left( {{f}_{{{{{\theta }'}}_{i}}}} \right)}
\end{equation}
where $\beta$ is the meta step size. It assembles the total sample batch of tasks ${{\mathcal{T}}_{i}}\sim p\left( \mathcal{T} \right)$ to update the model ${{f}_{{{{{\theta }'}}_{i}}}}$, which appearing Algorithm~\ref{alg:training} (line~\ref{line:meta}).

In the test session, we test the trained model using the test dataset, and we perform online decoding inference (summarized in Algorithm~\ref{alg:testing} ) using  decoder 1 that obtains global attention and compares it with the real value to obtain the reconstruction loss.

\begin{algorithm}[!t]
	\begin{algorithmic}[1]
		\Require
		\Statex Testing time series $\text{X}\in {{\mathbb{R}}^{T\times M}}$
		\Statex $\hat{\mathcal{W}}$ for testing sliding windows
		\Statex Split time series into the dataset $\hat{\mathcal{D}}={{\left\{ {{\mathcal{X}}_{d}} \right\}}_{1:D}},\mathcal{X}\in {{\mathbb{R}}^{\mathcal{W}\times M}}$
		\Statex Encoders ${{E}_{1}}$ and ${{E}_{2}}$, Decoders ${{D}_{1}}$
		\Statex Set hyperparameters $\text{W}$
		\State Well-trained parameters ${{\hat{\Theta }}_{e}},{{\hat{\Theta }}_{d}}$
		\State Randomly sample one batch from the dataset $\hat{\mathcal{D}}$
		\State \textbf{for} $t=1$ to $\hat{\mathcal{W}}$ \textbf{do}
		\State \hspace{8pt} ${{\hat{\mathcal{O}}}_{1}}\leftarrow \varphi _{\phi }^{{{D}_{1}}}\left( \phi \left( {{\mathcal{X}}_{:,t}};{{{\hat{\Theta }}}_{{{E}_{1}}}} \right);\text{W} \right)$ 
		\State \hspace{8pt} $s={{\left\| {{{\hat{\mathcal{O}}}}_{1}}-{{\mathcal{X}}_{t}} \right\|}_{2}}$ 
		\State \hspace{8pt} ${{y}_{i}}\left( {{s}_{i}}\ge \text{POT}\left( {{s}_{i}} \right) \right)$
		\State \hspace{8pt} $D=\underset{i}{\mathop{\vee }}\,{{y}_{i}}$ 
		\Ensure $D$;
	\end{algorithmic}
	\caption{Network Testing Procedure}
	\label{alg:testing}
\end{algorithm}

\subsection{Anomaly Detection and Diagnosis}
\label{sec:Anomaly Detection}
% algo, inference, anomaly score, POT, anomaly labels for each dimension (diagnosis), anomaly labels overall (detection).
% Add visualization of anomaly scores as well (maybe with tSNE?)
The online inference is performed sequentially for the input data of our current sliding window, and we will continuously obtain anomaly scores for each dimensional timestamp. We apply the POT~\cite{siffer2017anomaly} approach to select thresholds dynamically and automatically for each dimension. During offline training, the anomaly scores of each dimension will form a univariate time series $S=\left\{ {{S}_{1}},{{S}_{2}},...{{S}_{N}} \right\}$. Next, we set the threshold $th{{r}_{F}}$ based on the offline anomaly score combined with the Extreme Value Theory (EVT).

Essentially, this is a statistical method. Instead of manually setting thresholds and making any assumptions about the distribution, it uses "extreme value theory" following Generalized Pareto Distribution (GPD) to fit the data and determine the proper value-at-risk (label) to dynamically set	the threshold. The POT is the second theorem in EVT. In our experiments, I mainly focus on anomalies distributed at the high end. The GPD function is shown below:
\begin{equation}
	\bar{F}(s)=P(\left. S-thr>s \right|S>thr)\sim {{\left( 1+\frac{\gamma s}{\beta } \right)}^{-\frac{1}{\gamma }}}
\end{equation}
Similar to~\cite{siffer2017anomaly}, where $thr$ is the initial threshold of anomaly scores. The part that exceeds the threshold is denoted as $S-thr$, which follows the Generalized Pareto Distribution with parameters $\gamma ,\beta$ with high probability. we estimate the values of parameters $\hat{\gamma }$ and $\hat{\beta }$ by Maximum Likelihood Estimation(MLE). The final threshold $th{{r}_{F}}$ can be computed by the following formula:
\begin{equation}
	th{{r}_{F}}\simeq thr-\frac{{\hat{\beta }}}{{\hat{\gamma }}}\left( {{\left( \frac{qN}{{{N}_{thr}}} \right)}^{-\hat{\gamma }}}-1 \right)
\end{equation}
where $q$ is the probability of desire to observe $S>thr$, $N$ is the number of observations, and ${{N}_{thr}}$ is the number of peaks \textit{i.e.} the number of ${{S}_{i}}$ s.t. ${{S}_{i}}>thr$. Here are two parameters ($thr$) and ($q$) that can be set based on experience: low quantile (\textit{e.g.}, less than $7\%$ and $q$ (${{10}^{-4}}$)), seeing Appendix~\ref{app:hyp}.

\begin{figure}[]
	\centering %\setlength{\belowcaptionskip}{-10pt}
	\includegraphics[width=1.0\linewidth,height=1.0\linewidth]{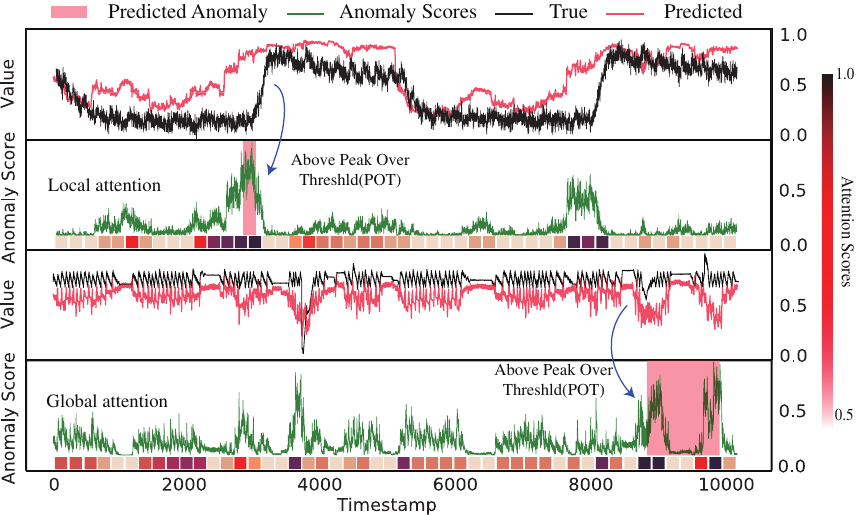}
	\caption{Detection effect of two dimensions in the WADI dataset.}
	\label{fig:test}
\end{figure}

\emph{\textbf{Impact of Dual Attention.}} Figure~\ref{fig:test} shows the detection effect of two dimensions in the WADI dataset with double attention. With global and local attention mechanisms, peaks and noise in the data receive extremely high attention. The model is designed with the dimensionality of the dataset as the amounts of heads for the multi-headed attention. We chose this number to enable the model to fully consider features across all dimensions, enhancing its ability to capture relationships and patterns in the data, which makes the attention scores of different dimensions also have some spatial correlation. Thus, our model is well able to detect anomalies in each dimension separately and specifically, using the context of the entire time series as priori.

\section{Experiments}
\label{sec:experiments}
In this chapter, we first describe the experimental dataset and performance metrics. In our baseline experiments, our model DTAAD is compared with the most popular and advanced methods currently available\footnote{We use publicly available code sources for most of the baselines. LSTM-NDT~\url{https://github.com/khundman/telemanom}, openGauss~\url{https://gitee.com/opengauss/openGauss-AI}, DAGMM \url{https://github.com/tnakae/DAGMM}, OmniAnomaly \url{https://github.com/NetManAIOps/OmniAnomaly}, MSCRED \url{https://github.com/7fantasysz/MSCRED}, MAD-GAN \url{https://github.com/LiDan456/MAD-GANs}. Other models were re-implemented by us.}.  Finally, the necessity of the components of our model is verified by ablation experiments.

We set the hyperparameters as described in the baseline models in their respective papers. We used the PyTorch-1.10.1~\cite{paszke2019pytorch} library to train all models. For more experimental details refer to Appendix~\ref{app:hyp}.

\begin{table}[t]
	\centering  %\renewcommand*{\arraystretch}{0.92}
	\caption{Dataset Information}
	\resizebox{\linewidth}{!}{
		\begin{tabular}{@{}llrrr@{}}
			\toprule
			Dataset & Dimensions & Train & Test & Anomalies ($\%$)\tabularnewline
			\midrule
			MSDS & 10 (1) & 146430 & 146430 & 5.37\tabularnewline
			SMD & 38 (4) & 708420 & 708420 & 4.16\tabularnewline
			WADI & 123 (1) & 1048571 & 172801 & 5.99\tabularnewline
			SWaT & 51 (1) & 496800 & 449919 & 11.98\tabularnewline
			MSL & 55 (3) & 58317 & 73729 & 10.72\tabularnewline
			SMAP & 25 (55) & 135183 & 427617 & 13.13\tabularnewline
			MBA & 2 (8) & 100000 & 100000 & 0.14\tabularnewline
			UCR & 1 (4) & 1600 & 5900 & 1.88\tabularnewline
			NAB & 1 (6) & 4033 & 4033 & 0.92\tabularnewline
			\bottomrule
	\end{tabular}}
	\label{tab:datasets}
\end{table} 
\raggedbottom

\subsection{Datasets}
\label{sec:datasets}
We adopt nine available datasets in our experiments (Seven of them are public data sets). We summarize their features in Table~\ref{tab:datasets}. Table~\ref{tab:datasets} illustrates the details of these datasets, including the name, the number of entities, the number of dimensions, the size of the train and test sets, and the anomaly rate for each test subset. For example, the MBA dataset has 8 entities, each with 2 dimensions.
\begin{enumerate}
	\item \textit{Numenta Anomaly Benchmark (NAB)}: This is a benchmark containing a real data stream with marked exceptions, containing multiple real data traces~\cite{numenta}. The current NAB data includes various kinds of sources, from social media to temperature sensors to server network utilization, and incorrectly tagged sequences of anomalies were removed from our experiments.
	\item \textit{HexagonML (UCR) dataset}: is a dataset consisting of multivariate time series~\cite{dau2019ucr} that was used in the KDD 2021 cup. We used only the portion of the dataset obtained from the real world.
	\item \textit{MIT-BIH Supraventricular Arrhythmia Database (MBA)}: This is the first generally available set of standard test materials for the evaluation of arrhythmia detectors, and it has been used in about 500 basic studies of cardiac dynamics~\cite{moody2001impact}. This is a popular large dataset.
	\item \textit{Soil Moisture Active Passive (SMAP) dataset}: This is a 25-dimensional public dataset collected by NASA~\cite{hundman2018detecting} that contains telemetry information anomaly data extracted from Anomalous Event Anomaly (ISA) reports from spacecraft monitoring systems.
	\item \textit{Mars Science Laboratory (MSL) dataset}: This is a SMAP-like dataset that includes actuator and sensor data from the Mars rover itself~\cite{hundman2018detecting}. We adopt only the three non-trivial ones (A4, C2, and T1).
	\item \textit{Secure Water Treatment (SWaT) dataset}: This is the collected data obtained from 51 sensors of the critical infrastructure in the continuously operating water treatment system~\cite{mathur2016swat} (water level, flow rate, \textit{etc}.).
	\item \textit{Water Distribution (WADI) dataset}: WADI is currently used for security analysis of water distribution networks and experimental evaluation of detection mechanisms for physical attacks and potential cyber~\cite{ahmed2017wadi}. Longer collection time for the dataset, including 14 days of normal scenarios and 2 days of attack scenarios. For the WADI dataset, we utilized zero values to fill in the missing values.
	\item \textit{Server Machine Dataset (SMD)}: This is a new 5-week-long dataset collected from a large Internet company~\cite{su2019robust}, and SMD is divided into two subsets of equal size as the training and test sets. Similar to MSL, we use the four non-trivial sequences in this dataset.
	\item \textit{Multi-Source Distributed System (MSDS) Dataset}: This is multi-source distributed system data for AI analysis, consisting of application logs, metrics, and distributed traces~\cite{nedelkoski2020multi}.
\end{enumerate}

\begin{table*}[!t]
	\centering %\renewcommand*{\arraystretch}{0.92}
	\caption{Performance comparison of DTAAD with baseline methods on the nine complete datasets. The P, R, AUC, and F1 represent the Precision, Recall, and Area under the ROC curve and F1 scores respectively. The best AUC and F1 scores are highlighted in bold.} % AUC*: AUC with 20\% training data, F1*: F1 score with 20\% training data. 
	\resizebox{\linewidth}{!}{
	\begin{tabular}{llcccccccccccccc}
		\toprule                                       
		\multirow{2}{*}{Method} &  & \multicolumn{4}{c}{NAB} &  & \multicolumn{4}{c}{UCR} &  & \multicolumn{4}{c}{MBA}\tabularnewline
		\cmidrule{3-16} 
		&  & P & R & AUC & F1 &  & P & R & AUC & F1 &  & P & R & AUC & F1\tabularnewline
		\midrule 
		LSTM-NDT &  & 0.5333 & 0.6667 & 0.8316 & 0.5926	&  & 0.5320	& 0.8293 & 0.9780 & 0.5230	& & 0.9206  & 0.9717 & 0.9779 & 0.9455\tabularnewline
		DAGMM &  & 0.7621 & 0.7291 & 0.8571 & 0.7452 &  & 0.5336 & 0.9717 & 0.9915 & 0.6889 &  & 0.9474 & 0.9900 & 0.9857 & 0.9682\tabularnewline
		OmniAnomaly &  & 0.8420 & 0.6666 & 0.8329 & 0.7441 &  & 0.8345 & 0.9999 & 0.9980 & 0.9097 &  & 0.8560 & 1.0000 & 0.9569 & 0.9224\tabularnewline
		MSCRED &  & 0.8521 & 0.6700 & 0.8400 & 0.7501 &  & 0.5440 & 0.9717 & 0.9919 & 0.6975 &  & 0.9271 & 1.0000 & 0.9798 & 0.9622\tabularnewline
		MAD-GAN &  & 0.8665 & 0.7011 & 0.8477 & 0.7751 &  & 0.8537 & 0.9890 & 0.9983 & 0.9164 &  & 0.9395 & 1.0000 & 0.9835 & 0.9688\tabularnewline
		USAD &  & 0.8421 & 0.6667 & 0.8332 & 0.7443 &  & 0.8953 & 1.0000 & 0.9990 & 0.8953 &  & 0.8954 & 0.9990 & 0.9702 & 0.9444\tabularnewline
		MTAD-GAT &  & 0.8422 & 0.7273 & 0.8222 & 0.7803 &  & 0.7813 & 0.9973 & 0.9979 & 0.8762 &  & 0.9019 & 1.0000 & 0.9720 & 0.9483\tabularnewline
		CAE-M &  & 0.7919 & 0.8020 & 0.8020 & 0.7969 &  & 0.6982 & 1.0000 & 0.9958 & 0.8223 &  & 0.8443 & 0.9998 & 0.9662 & 0.9155\tabularnewline
		GDN &  & 0.8130 & 0.7873 & 0.8543 & 0.7999 &  & 0.6895 & 0.9989 & 0.9960 & 0.8159 &  & 0.8833 & 0.9893 & 0.9529 & 0.9333\tabularnewline
		TranAD &  & 0.8889 & 0.9892 & 0.9541 & 0.9364 &  & 0.9407 & 1.0000 & \textbf{0.9994} & \textbf{0.9694} &  & 0.9576 & 1.0000 & 0.9886 & 0.9783\tabularnewline
		\textbf{DTAAD} &  & 0.8889 & 0.9999 & \textbf{0.9996} & \textbf{0.9412} &  & 0.8880 & 1.0000 & 0.9988 & 0.9407 &  & 0.9608 & 1.0000 & \textbf{0.9896} & \textbf{0.9800}\tabularnewline
		\midrule 
		\multirow{2}{*}{Method} &  & \multicolumn{4}{c}{SMAP} &  & \multicolumn{4}{c}{MSL} &  & \multicolumn{4}{c}{SWaT}\tabularnewline
		\cmidrule{3-16} 
		&  & P & R & AUC & F1 &  & P & R & AUC & F1 &  & P & R & AUC & F1\tabularnewline    
		\midrule 
		LSTM-NDT &  & 0.8522 & 0.7325 & 0.8601 & 0.7877 &  & 0.6287 & 1.0000 & 0.9531 & 0.7720 &  & 0.7777 & 0.5108 & 0.7139 & 0.6166\tabularnewline
		DAGMM &  & 0.8068 & 0.9890 & 0.9884 & 0.8887 &  & 0.7362 & 1.0000 & 0.9715 & 0.8481 &  & 0.9932 & 0.6878 & 0.8435 & 0.8127\tabularnewline
		OmniAnomaly &  & 0.8129 & 0.9418 & 0.9888 & 0.8727 &  & 0.7847 & 0.9923 & 0.9781 & 0.8764 &  & 0.9781 & 0.6956 & 0.8466 & 0.8130\tabularnewline
		MSCRED &  & 0.8174 & 0.9215 & 0.9820 & 0.8663 &  & 0.8911 & 0.9861 & 0.9806 & 0.9362 &  & 0.9991 & 0.6769 & 0.8432 & 0.8071\tabularnewline
		MAD-GAN &  & 0.8156 & 0.9215 & 0.9890 & 0.8653 &  & 0.8515 & 0.9929 & 0.9861 & 0.9168 &  & 0.9592 & 0.6956 & 0.8462 & 0.8064\tabularnewline
		USAD &  & 0.7481 & 0.9628 & 0.9890 & 0.8419 &  & 0.7949 & 0.9912 & 0.9795 & 0.8822 &  & 0.9977 & 0.6879 & 0.8460 & 0.8143\tabularnewline
		MTAD-GAT &  & 0.7992 & 0.9992 & 0.9846 & 0.8882 &  & 0.7918 & 0.9825 & 0.9890 & 0.8769 &  & 0.9719 & 0.6958 & 0.8465 & 0.8110\tabularnewline
		CAE-M &  & 0.8194 & 0.9568 & 0.9902 & 0.8828 &  & 0.7752 & 1.0000 & 0.9904 & 0.8734 &  & 0.9698 & 0.6958 & 0.8465 & 0.8102\tabularnewline
		GDN &  & 0.7481 & 0.9892 & 0.9865 & 0.8519 &  & 0.9309 & 0.9893 & 0.9815 & 0.9592 &  & 0.9698 & 0.6958 & 0.8463 & 0.8102\tabularnewline
		TranAD &  & 0.8104 & 0.9998 & 0.9887 & 0.8953 &  & 0.9037 & 0.9999 & 0.9915 & 0.9493 &  & 0.9977 & 0.6879 & 0.8438 & \textbf{0.8143}\tabularnewline
		\textbf{DTAAD} &  & 0.8220 & 0.9999 & \textbf{0.9911} & \textbf{0.9023} &  & 0.9038 & 0.9999 & \textbf{0.9918} & \textbf{0.9495} &  & 0.9697 & 0.6957 & \textbf{0.8462} & 0.8101\tabularnewline
		\midrule 
		\multirow{2}{*}{Method} &  & \multicolumn{4}{c}{WADI} &  & \multicolumn{4}{c}{SMD} &  & \multicolumn{4}{c}{MSDS}\tabularnewline
		\cmidrule{3-16} 
		&  & P & R & AUC & F1 &  & P & R & AUC & F1 &  & P & R & AUC & F1\tabularnewline
		\midrule 
		LSTM-NDT &  & 0.0137 & 0.7822 & 0.6720 & 0.0270 &  & 0.9735 & 0.8439 & 0.9670 & 0.9041 &  & 0.9998 & 0.8011 & 0.8012 & 0.8896\tabularnewline
		DAGMM &  & 0.0759 & 0.9980 & 0.8562 & 0.1411 &  & 0.9102 & 0.9913 & 0.9953 & 0.9490 &  & 0.9890 & 0.8025 & 0.9012 & 0.8861\tabularnewline
		OmniAnomaly &  & 0.3157 & 0.6540 & 0.8197 & 0.4259 &  & 0.8880 & 0.9984 & 0.9945 & 0.9400 &  & 0.9998 & 0.7963 & 0.8981 & 0.8867\tabularnewline
		MSCRED &  & 0.2512 & 0.7318 & 0.8411 & 0.3740 &  & 0.7275 & 0.9973 & 0.9920 & 0.8413 &  & 0.9998 & 0.7982 & 0.8942 & 0.8878\tabularnewline
		MAD-GAN &  & 0.2232 & 0.9123 & 0.8025 & 0.3587 &  & 0.9990 & 0.8439 & 0.9932 & 0.9149 &  & 0.9981 & 0.6106 & 0.8053 & 0.7578\tabularnewline
		USAD &  & 0.1874 & 0.8297 & 0.8724 & 0.3057 &  & 0.9061 & 0.9975 & 0.9934 & 0.9496 &  & 0.9913 & 0.7960 & 0.8980 & 0.8829\tabularnewline
		MTAD-GAT &  & 0.2819 & 0.8013 & 0.8822 & 0.4170 &  & 0.8211 & 0.9216 & 0.9922 & 0.8684 &  & 0.9920 & 0.7965 & 0.8983 & 0.8835\tabularnewline
		CAE-M &  & 0.2783 & 0.7917 & 0.8727 & 0.4118 &  & 0.9081 & 0.9670 & 0.9782 & 0.9368 &  & 0.9909 & 0.8440 & 0.9014 & 0.9115\tabularnewline
		GDN &  & 0.2913 & 0.7932 & 0.8778 & 0.4261 &  & 0.7171 & 0.9975 & 0.9925 & 0.8343 &  & 0.9990 & 0.8027 & 0.9106 & 0.8900\tabularnewline
		TranAD &  & 0.3959 & 0.8295 & \textbf{0.8998} & 0.5360 &  & 0.9051 & 0.9973 & \textbf{0.9933} & \textbf{0.9490} &  & 0.9998 & 0.8625 & 0.9012 & 0.8904\tabularnewline
		\textbf{DTAAD} &  & 0.9017 & 0.3910 & 0.6950 & \textbf{0.5455} &  & 0.8463 & 0.9974 & 0.9892 & 0.9147 &  & 0.9999 & 0.8026 & \textbf{0.9013} & \textbf{0.8905}\tabularnewline
		\bottomrule 
	\end{tabular}}
	\label{tab:detection}
\end{table*}

\subsection{Performance Metrics}
We adopt Precision, Recall, F1-Score and AUC value (area under the ROC curve) as metrics to evaluate the DTAAD and baseline models.
\begin{equation}
	Precision=\frac{TP}{TP+FP}
\end{equation}
\begin{equation}
	Recall=\frac{TP}{TP+FN}
\end{equation}
\begin{equation}
	\text{ }F1=\frac{2\times Precision\times Recall}{Precision+Recall}
\end{equation}

To validate the model performance with limited data, we also set up to test the above metrics on all model training with only $20\%$ training data (AUC* and F1*). We also introduced two other metrics to measure performance: $\mathrm{HitRate@P\%}$ is to rank the anomaly scores and determine how many true dimensions are hit by the top-ranked priority~\cite{su2019robust}. For instance, if at timestamp $t$, 2 dimensions are labeled anomalous in the ground truth, $\mathrm{HitRate@100\%}$ would consider top 2 dimensions and $\mathrm{HitRate@150\%}$ would consider 3 dimensions. $\mathrm{NDCG@P\%}$ is the Normalized Discounted Cumulative Gain (NDCG)~\cite{su2019robust}. The main principle of NDCG is to compute the anomaly scores, and then apply a discount factor to diminish the weight of lower ranks. This implies that higher-ranked projects contribute more to loss, and their normalized value eliminates inter-dataset dimensionality differences. Finally, the loss value obtained is divided by the theoretical best-ranked loss value. $\mathrm{NDCG@P\%}$ and $\mathrm{HitRate@P\%}$ priorities are considered in the same way.

\begin{table}[!t]
	\centering %\renewcommand*{\arraystretch}{0.92}
	\caption{Performance comparison of DTAAD with baseline methods with $20\%$ of the training dataset. The AUC* and F1* represent the AUC with $20\%$ training data and F1 score with $20\%$ training data. The best AUC* and F1* scores are highlighted in bold.} 
	\resizebox{\linewidth}{!}{
		\begin{tabular}{@{}lcccccc@{}}
			\toprule 
			\multirow{2}{*}{Method} & \multicolumn{2}{c}{NAB} & \multicolumn{2}{c}{UCR} & \multicolumn{2}{c}{MBA}\tabularnewline
			\cmidrule{2-7} 
			& AUC{*} & F1{*} & AUC{*} & F1{*} & AUC{*} & F1{*}\tabularnewline
			\midrule 
			LSTM-NDT & 0.8012 & 0.6211 & 0.8912 & 0.5197 & 0.9616 & 0.9281\tabularnewline
			DAGMM & 0.7826 & 0.6124 & 0.9811 & 0.5717 & 0.9670 & 0.9395\tabularnewline
			OmniAnomaly & 0.8128 & 0.6712 & 0.9727 & 0.7917 & 0.9406 & 0.9216\tabularnewline
			MSCRED & 0.8298 & 0.7012 & 0.9636 & 0.4928 & 0.9498 & 0.9107\tabularnewline
			MAD-GAN & 0.8193 & 0.7108 & 0.9958 & 0.8215 & 0.9549 & 0.9191\tabularnewline
			USAD & 0.7268 & 0.6782 & 0.9968 & 0.8539 & 0.9698 & 0.9426\tabularnewline
			MTAD-GAT & 0.6957 & 0.7012 & 0.9975 & 0.8672 & 0.9689 & 0.9426\tabularnewline
			CAE-M & 0.7313 & 0.7127 & 0.9927 & 0.7526 & 0.9617 & 0.9003\tabularnewline
			GDN & 0.8300 & 0.7014 & 0.9938 & 0.8030 & 0.9672 & 0.9317\tabularnewline
			TranAD & 0.9216 & 0.8420 & 0.9983 & 0.9211 & 0.9946 & 0.9897\tabularnewline
			\textbf{DTAAD} & \textbf{0.9330} & \textbf{0.9057} & \textbf{0.9984} & \textbf{0.9220} & \textbf{0.9955} & \textbf{0.9912}\tabularnewline
			\midrule 
			\multirow{2}{*}{Method} & \multicolumn{2}{c}{SMAP} & \multicolumn{2}{c}{MSL} & \multicolumn{2}{c}{SWaT}\tabularnewline
			\cmidrule{2-7} 
			& AUC{*} & F1{*} & AUC{*} & F1{*} & AUC{*} & F1{*}\tabularnewline
			\midrule 
			LSTM-NDT & 0.7006 & 0.5417 & 0.9519 & 0.7607 & 0.6689 & 0.4144\tabularnewline
			DAGMM & 0.9880 & 0.8368 & 0.9605 & 0.8009 & 0.8420 & 0.8000\tabularnewline
			OmniAnomaly & 0.9878 & 0.8130 & 0.9702 & 0.8423 & 0.8318 & 0.7432\tabularnewline
			MSCRED & 0.9810 & 0.8049 & 0.9796 & 0.8231 & 0.8384 & 0.7921\tabularnewline
			MAD-GAN & 0.9876 & 0.8467 & 0.9648 & 0.8189 & 0.8455 & 0.8011\tabularnewline
			USAD & 0.9884 & 0.8380 & 0.9650 & 0.8191 & 0.8439 & 0.8088\tabularnewline
			MTAD-GAT & 0.9815 & 0.8226 & 0.9783 & 0.8025 & 0.8460 & 0.8080\tabularnewline
			CAE-M & 0.9893 & 0.8313 & 0.9837 & 0.7304 & 0.8459 & 0.7842\tabularnewline
			GDN & 0.9888 & 0.8412 & 0.9415 & 0.8960 & 0.8391 & 0.8073\tabularnewline
			TranAD & 0.9884 & 0.8936 & 0.9856 & 0.9171 & 0.8461 & \textbf{0.8093}\tabularnewline
			\textbf{DTAAD} & \textbf{0.9894} & \textbf{0.8996} & \textbf{0.9864} & \textbf{0.9212} & \textbf{0.8460} & 0.8087\tabularnewline
			\midrule 
			\multirow{2}{*}{Method} & \multicolumn{2}{c}{WADI} & \multicolumn{2}{c}{SMD} & \multicolumn{2}{c}{MSDS}\tabularnewline
			\cmidrule{2-7} 
			& AUC{*} & F1{*} & AUC{*} & F1{*} & AUC{*} & F1{*}\tabularnewline
			\midrule 
			LSTM-NDT & 0.6636 & 0.0000 & 0.9562 & 0.6753 & 0.7812 & 0.7911\tabularnewline
			DAGMM & 0.6496 & 0.0629 & 0.9844 & 0.8985 & 0.7762 & 0.8388\tabularnewline
			OmniAnomaly & \textbf{0.7912} & \textbf{0.1016} & 0.9857 & \textbf{0.9351} & 0.5612 & 0.8388\tabularnewline
			MSCRED & 0.6028 & 0.0412 & 0.9767 & 0.8003 & 0.7715 & 0.8282\tabularnewline
			MAD-GAN & 0.5382 & 0.0936 & 0.8634 & 0.9317 & 0.5001 & 0.7389\tabularnewline
			USAD & 0.7012 & 0.0734 & 0.9855 & 0.9214 & 0.7614 & 0.8390\tabularnewline
			MTAD-GAT & 0.6268 & 0.0521 & 0.9799 & 0.6662 & 0.6123 & 0.8249\tabularnewline
			CAE-M & 0.6110 & 0.0782 & 0.9570 & 0.9319 & 0.6002 & 0.8390\tabularnewline
			GDN & 0.6122 & 0.0413 & 0.9812 & 0.7108 & 0.6820 & 0.8390\tabularnewline
			TranAD & 0.6852 & 0.0698 & 0.9847 & 0.8794 & 0.8112 & 0.8389\tabularnewline
			\textbf{DTAAD} & 0.7818 & 0.0977 & \textbf{0.9866} & 0.8941 & \textbf{0.8115} & \textbf{0.8390}\tabularnewline
			\bottomrule 
	\end{tabular}}
	\label{tab:detection20}
\end{table}

\subsection{Results and Analysis}

Tables~\ref{tab:detection} and~\ref{tab:detection20} show the P, R, AUC, F1, AUC*, and F1* scores of DTAAD and baseline methods on nine datasets. Our model outperforms most of the baseline, both on the full dataset and on the training dataset with only $20\%$. Specifically, on the total dataset (\textit{i.e.}, the union of these nine datasets) compared to the mean of all comparison baselines (\textit{i.e.}, the arithmetic mean of the DTAAD model scores for the nine datasets is compared to the arithmetic mean of the respective scores of all other models when summed and then averaged), the F1 score for DTAAD improved by $8.38\%$, F1* score by $8.97\%$, AUC score by $1\%$, and AUC* score by $9.98\%$.

On the complete training dataset, the DTAAD model outperforms most baseline models except for TranAD. For UCR and SMD datasets, TranAD achieves the highest F1 and AUC scores. On SWaT, TranAD achieves the highest F1 score (0.8143), while on WADI it attains the highest AUC score (0.8998). Similarly, on $20\%$ of the training dataset, the DTAAD model outperforms the baseline models on all datasets except for WADI, where OmniAnomaly performs well and achieves the highest AUC* score (0.7912) and the highest F1* score (0.1016). In summary, our model outperforms most of the baseline, both on the full dataset and on the training dataset with only $20\%$. 

Specifically, DAGMM models perform well on short datasets such as NAB, UCR, MBA, and SMAP, but their detection capabilities significantly diminish when dealing with longer sequences. This is because it does not use the sliding sequence window. DTAAD's global TCN can fully map long sequence information and is therefore capable of detecting anomalies. Other models such as OmniAnomaly, CAE-M, and MSCRED employ sequential input but this can lead to near-normal trends being ignored due to the accumulation of long-term information. In contrast, DTAAD uses a ratio feedback mechanism to amplify anomalies. More recent models like USAD, MTAD-GAT, and GDN also only use local windows as inputs for the reconstruction of the prediction sequence. Both TranAD and DTAAD utilize position embeddings, which undoubtedly enhance the ability to capture long-term trends in sequences. However, due to the inherent limitations of Transformer, which can only see the local context window, TranAD fails to fully compensate for this. Moreover, although OmniAnomaly performs well on the $20\%$ WADI dataset, this is due to the specific reconstruction design of the model being better suited for fewer datasets with higher noise levels. However, DTAAD only slightly trails OmniAnomaly in terms of F1* and AUC* on the WADI dataset. On the other hand, all models perform relatively poorly on the full WADI dataset due to its massive sequence length and complex dimensions. However, it becomes clear that DTAAD outperforms all other models in terms of detection accuracy in the WADI dataset.

\begin{table*}[]
	\centering %\renewcommand*{\arraystretch}{0.92}
	\caption{Comparison of diagnosing Performance.}
	\begin{tabular}{llccccccccc}
		\toprule 
		\multirow{2}{*}{Method} &  & \multicolumn{4}{c}{SMD} &  & \multicolumn{4}{c}{MSDS}\tabularnewline
		\cmidrule{3-11}
		&  & H@100$\%$ & H@150$\%$ & N@100$\%$ & N@150$\%$ &  & H@100$\%$ & H@150$\%$ & N@100$\%$ & N@150$\%$\tabularnewline
		\midrule 
		LSTM-NDT &  & 0.1920 & 0.2951 & 0.2586 & 0.3175 &  & 0.3166 & 0.4551 & 0.2966 & 0.3780\tabularnewline
		DAGMM &  & 0.2617 & 0.4333 & 0.3153 & 0.4151 &  & 0.4627 & 0.6021 & 0.5169 & 0.5658\tabularnewline
		OmniAnomaly &  & 0.2839 & 0.4365 & 0.3338 & 0.4231 &  & 0.3967 & 0.5652 & 0.4545 & 0.5125\tabularnewline
		MSCRED &  & 0.2322 & 0.3469 & 0.2297 & 0.2962 &  & 0.3672 & 0.5180 & 0.4609 & 0.5164\tabularnewline
		MAD-GAN &  & 0.3856 & 0.5589 & 0.4277 & 0.5292 &  & 0.4030 & 0.5785 & 0.4681 & 0.5522\tabularnewline
		USAD &  & 0.3095 & 0.4769 & 0.3534 & 0.4515 &  & 0.4325 & 0.6005 & \textbf{0.5179} & 0.5711\tabularnewline
		MTAD-GAT &  & \textbf{0.4912} & 0.5885 & 0.4126 & 0.4822 &  & 0.3493 &0.4777 & 0.3759 & 0.4530\tabularnewline
		CAE-M &  & 0.2530 & 0.4171 & 0.2047 & 0.3010 &  & 0.4107 & 0.4707 & 0.5174 & 0.5528\tabularnewline
		GDN &  & 0.2276 & 0.3382 & 0.2921 & 0.3570 &  & 0.3143 & 0.4386 & 0.2980 & 0.3724\tabularnewline
		TranAD &  & 0.3710 & 0.5509 & 0.4106 & 0.5204 &  & 0.4392 & 0.5797 & 0.4688 & 0.5528\tabularnewline
		\textbf{DTAAD} &  & 0.4087 & \textbf{0.6235} & \textbf{0.4298} & \textbf{0.5565} &  & \textbf{0.4636} & \textbf{0.6034} & 0.4887 & \textbf{0.5718}\tabularnewline
		\bottomrule 
	\end{tabular}
	\label{tab:diagnosis}
\end{table*}

\begin{table}[]
	\centering 
	\Large
	\renewcommand*{\arraystretch}{1.65}
	\caption{Comparison of training and inference time per epoch in seconds.}
	\resizebox{.475\textwidth}{!}{  % Here 1/2
		\begin{tabular}{lllcccccccc}
			\toprule 
			Method &  & NAB & UCR & MBA & SMAP & MSL & SWaT & WADI & SMD & MSDS\tabularnewline
			\midrule 
			LSTM-NDT &  & 2.10 & 1.74 & 5.56 & 5.52 & 5.25 & 5.29 & 59.42 & 74.63 & 72.22\tabularnewline
			DAGMM &  & 4.38 & 4.16 & 14.92 & 3.81 & 3.28 & 3.70 & 80.98 & 40.87 & 37.51\tabularnewline
			OmniAnomaly &  & 7.65 & 5.59 & 21.97 & 5.41 & 4.26 & 5.68 & 96.47 & 55.39 & 55.42\tabularnewline
			MSCRED &  & 51.77 & 52.49 & 118.43 & 3.23 & 6.69 & 36.73 & 269.81 & 47.53 & 21.93\tabularnewline
			MAD-GAN &  & 7.96 & 5.14 & 30.06 & 5.90 & 5.25 & 5.56 & 104.63 & 62.96 & 57.05\tabularnewline
			USAD &  & 6.24 & 4.22 & 24.17 & 4.73 & 4.24 & 4.54 & 109.52 & 50.19 & 46.56\tabularnewline
			MTAD-GAT &  & 29.01 & 19.42 & 46.62 & 203.01 & 257.48 & 20.78 & 1962.43 & 1312.82 & 260.82\tabularnewline
			CAE-M &  & 4.50 & 3.88 & 13.49 & 37.47 & 115.19 & 8.25 & 1105.12 & 620.42 & 110.57\tabularnewline
			GDN &  & 16.77 & 11.76 & 31.80 & 12.47 & 19.34 & 11.88 & 812.61 & 161.99 & 117.07\tabularnewline
			\textbf{DTAAD} &  & \textbf{0.51} & \textbf{0.23} & \textbf{0.97} & \textbf{1.47} & \textbf{2.70} & \textbf{0.37} & \textbf{52.33} & \textbf{24.47} & \textbf{6.38}\tabularnewline
			\bottomrule 
	\end{tabular}}  % Here 2/2
	\label{tab:overhead}
\end{table} 

Table~\ref{tab:diagnosis} shows the diagnostic performance where H and N refer to HitRate and NDCG (with complete data). Table~\ref{tab:overhead} shows the average training and inference time in seconds per epoch for all models on each dataset (we conducted model inferencing immediately after each epoch of training), calculated on the same machine and GPU with the same random seeds. Table~\ref{tab:diagnosis} illustrates that DTAAD is capable of detecting abnormalities in the range of $40.87\%$-$62.35\%$ and that DTAAD is able to increase the diagnostic scores of SMD and MSD by $7.5\%$ and $18\%$, respectively, compared to the baseline method. Table~\ref{tab:overhead} shows that the training and inference time of DTAAD is $62\%$-$99\%$ below the baseline methods (Given that TranAD's model also belongs to a very lightweight design, its model training and inference time does not significantly differ from the model training time of this paper's model, thus we have not included it in Table~\ref{tab:overhead} for comparison). Instead of sequential inference over a sliding window, we use the Transformer's positional encoding to take the complete sequence as input and combine a single Transformer encoding layer with a lightweight design of a Dual TCN. This is a strong indication of the lightweight nature of our model and the advantages of having positional encoding.

\subsection{Ablation Experiment}
%\subsection{Robustness w.r.t. Anomaly Contamination}
\begin{table}[]
	\centering %\renewcommand*{\arraystretch}{0.92}
	\caption{Ablation Study - F1* scores and AUC* for DTAAD and its ablated versions.}
	\resizebox{\linewidth}{!}{
		\begin{tabular}{@{}lcccccc@{}}
			\toprule 
			\multirow{2}{*}{Method} & \multicolumn{2}{c}{NAB} & \multicolumn{2}{c}{UCR} & \multicolumn{2}{c}{MBA}\tabularnewline
			\cmidrule{2-7} 
			& AUC{*} & F1{*} & AUC{*} & F1{*} & AUC{*} & F1{*}\tabularnewline
			\midrule 
			\textbf{DTAAD} & \textbf{0.9330} & \textbf{0.9057} & \textbf{0.9984} & \textbf{0.9220} & \textbf{0.9955} & \textbf{0.9912}\tabularnewline
			\hline 
			\usym{1F5F7} Tcn\_Local & 0.9325 & 0.9050 & 0.9980 & 0.9188 & 0.9952 & 0.9910\tabularnewline
			\usym{1F5F7} Tcn\_Global & 0.9325 & 0.8987 & 0.9980 & 0.9084 & 0.9952 & 0.9902\tabularnewline
			\usym{1F5F7} Callback & 0.9325 & 0.9055 & 0.9980 & 0.9195 & 0.9950 & 0.9903\tabularnewline
			\usym{1F5F7} Transformer & 0.9325 & 0.9050 & 0.9980 & 0.9188 & 0.9926 & 0.9858\tabularnewline
			\midrule 
			\multirow{2}{*}{Method} & \multicolumn{2}{c}{SMAP} & \multicolumn{2}{c}{MSL} & \multicolumn{2}{c}{SWaT}\tabularnewline
			\cmidrule{2-7} 
			& AUC{*} & F1{*} & AUC{*} & F1{*} & AUC{*} & F1{*}\tabularnewline
			\midrule 
			\textbf{DTAAD} & \textbf{0.9892} & \textbf{0.8985} & \textbf{0.9864} & \textbf{0.9212} & \textbf{0.8460} & \textbf{0.8088}\tabularnewline
			\hline 
			\usym{1F5F7} Tcn\_Local & 0.9869 & 0.8805 & 0.9196 & 0.6638 & 0.8459 & 0.8086\tabularnewline
			\usym{1F5F7} Tcn\_Global & 0.9870 & 0.8815 & 0.9316 & 0.6989 & 0.8459 & 0.8086\tabularnewline
			\usym{1F5F7} Callback & 0.9854 & 0.8688 & 0.9688 & 0.8356 & 0.8446 & 0.8000\tabularnewline
			\usym{1F5F7} Transformer & 0.9853 & 0.8682 & 0.9700 & 0.8412 & 0.8459 & 0.8086\tabularnewline
			\midrule 
			\multirow{2}{*}{Method} & \multicolumn{2}{c}{WADI} & \multicolumn{2}{c}{SMD} & \multicolumn{2}{c}{MSDS}\tabularnewline
			\cmidrule{2-7} 
			& AUC{*} & F1{*} & AUC{*} & F1{*} & AUC{*} & F1{*}\tabularnewline
			\midrule 
			\textbf{DTAAD} & \textbf{0.7818} & \textbf{0.0977} & \textbf{0.9866} & \textbf{0.8941} & \textbf{0.8115} & \textbf{0.8390}\tabularnewline
			\hline 
			\usym{1F5F7} Tcn\_Local & 0.7397 & 0.0832 & 0.9794 & 0.8417 & 0.8110 & 0.8389\tabularnewline
			\usym{1F5F7} Tcn\_Global & 0.7407 & 0.0835 & 0.9828 & 0.8654 & 0.5000 & 0.8389\tabularnewline
			\usym{1F5F7} Callback & 0.7767 & 0.0957 & 0.9860 & 0.8940 & 0.8112 & 0.8388\tabularnewline
			\usym{1F5F7} Transformer & 0.7762 & 0.0951 & 0.9862 & 0.8920 & 0.8110 & 0.8388\tabularnewline
			\bottomrule 
	\end{tabular}}
	\label{tab:ablation}
\end{table}

Table~\ref{tab:ablation} shows the necessity of each component of the DTAAD model and provides the following findings:
\begin{itemize}
	\item The complete DTAAD model outperforms all non-complete models on a $20\%$ training dataset, especially on larger datasets (\textit{e.g.}, WADI\&SMD), with significant improvements. This reflects the necessity of all components.
	\item Of all the model components, TCN\_Global and TCN\_Local design components have the most significant performance improvement. Therefore, removing these two components resulted in the biggest drop in F1* scores and AUC* for the majority of the datasets. This decline is even more obvious for the WADI and SMD datasets (In particular, the full dataset which is not shown here), proving the necessity of TCN\_Global for large datasets.
\end{itemize}

\section{Conclusion}
\label{sec:analyses}
In this article, we design an anomaly detection model DTAAD based on Transformer and Dual TCN, which allows for rapid and specific anomaly detection and diagnosis of multivariate time series. The codec in Transformer allows for fast modeling in an experimental study, DTAAD largely outperformed state-of-the-art on nine benchmark datasets, improving F1 and F1* scores by $8.38\%$ and $8.97\%$ for complete and small datasets, respectively, and providing a promising solution for unsupervised anomaly detection on high-dimensional data. Our model as a whole combines AR and AE, with the addition of the Transformer to capture the information of long sequences. Designing causal convolution and dilated convolution as local TCN and global TCN, introducing a feedback mechanism, loss ratio to improve detection accuracy and expand association differences. 

Our proposed DTAAD is a well-generalized model that can be suitable for various devices. It can achieve this goal with up to $99\%$ less training time than baseline methods, making DTAAD ideal for modern industrial and embedded systems requiring accurate, fast anomaly prediction.

% if have a single appendix:
%\appendix[Proof of the Zonklar Equations]
% or
%\appendix  % for no appendix heading
% do not use \section anymore after \appendix, only \section*
% is possibly needed

% use appendices with more than one appendix
% then use \section to start each appendix
% you must declare a \section before using any
% \subsection or using \label (\appendices by itself
% starts a section numbered zero.)
%

%%\appendices
%%\section{Proof of the First Zonklar Equation}
%%Appendix one text goes here.

% you can choose not to have a title for an appendix
% if you want by leaving the argument blank
%%\section{}
%%Appendix two text goes here.
\appendices
\section{HYPER-PARAMETERS}
\label{app:hyp}

\begin{figure}
	\centering %\setlength{\belowcaptionskip}{-10pt}
	\includegraphics[width=\linewidth,,height=1.1\linewidth]{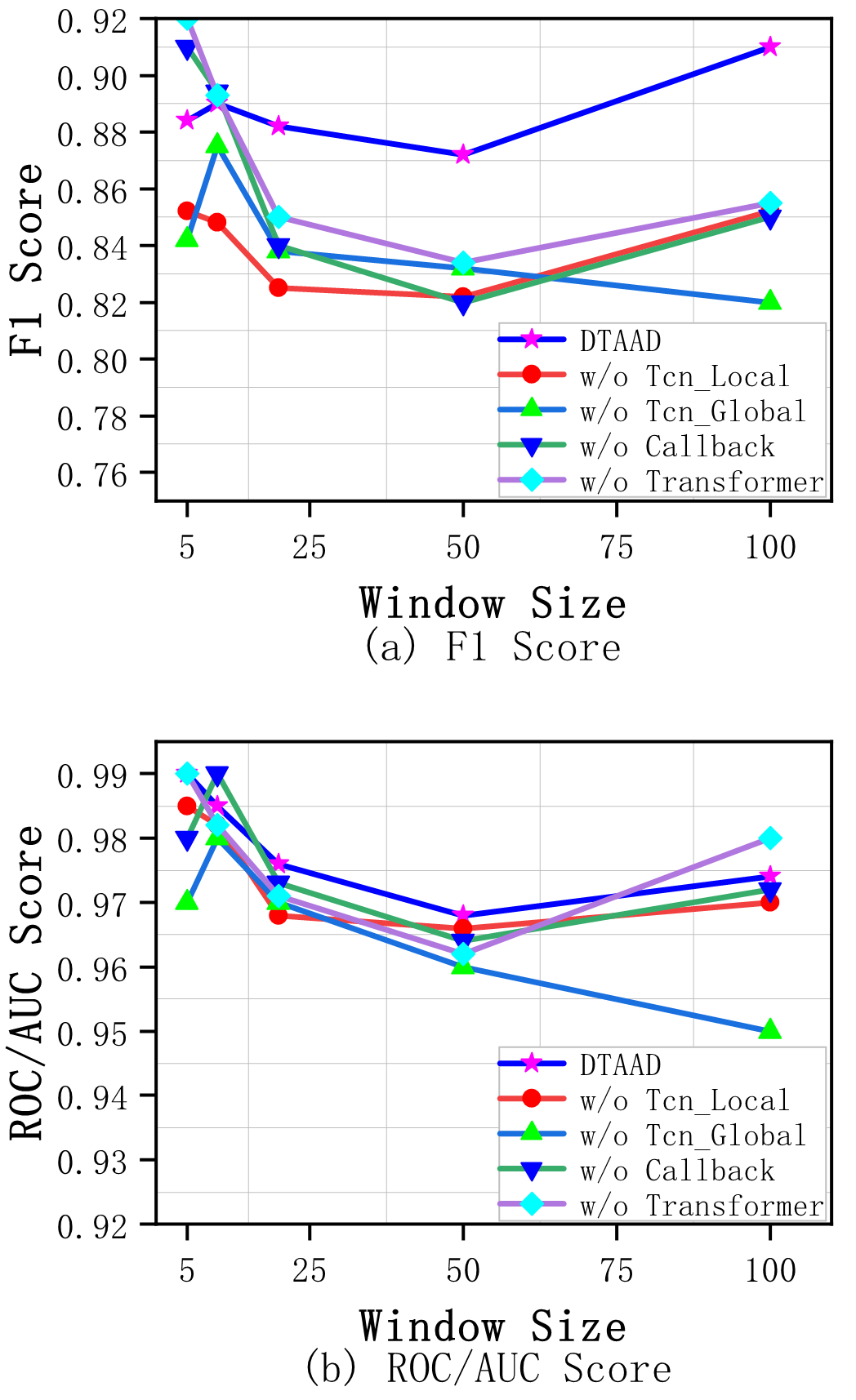}
	\caption{F1 score and ROC/AUC score with window size.}
	\label{fig:window size}
\end{figure}

\begin{figure}
	\centering %\setlength{\belowcaptionskip}{-10pt}
	\includegraphics[width=\linewidth]{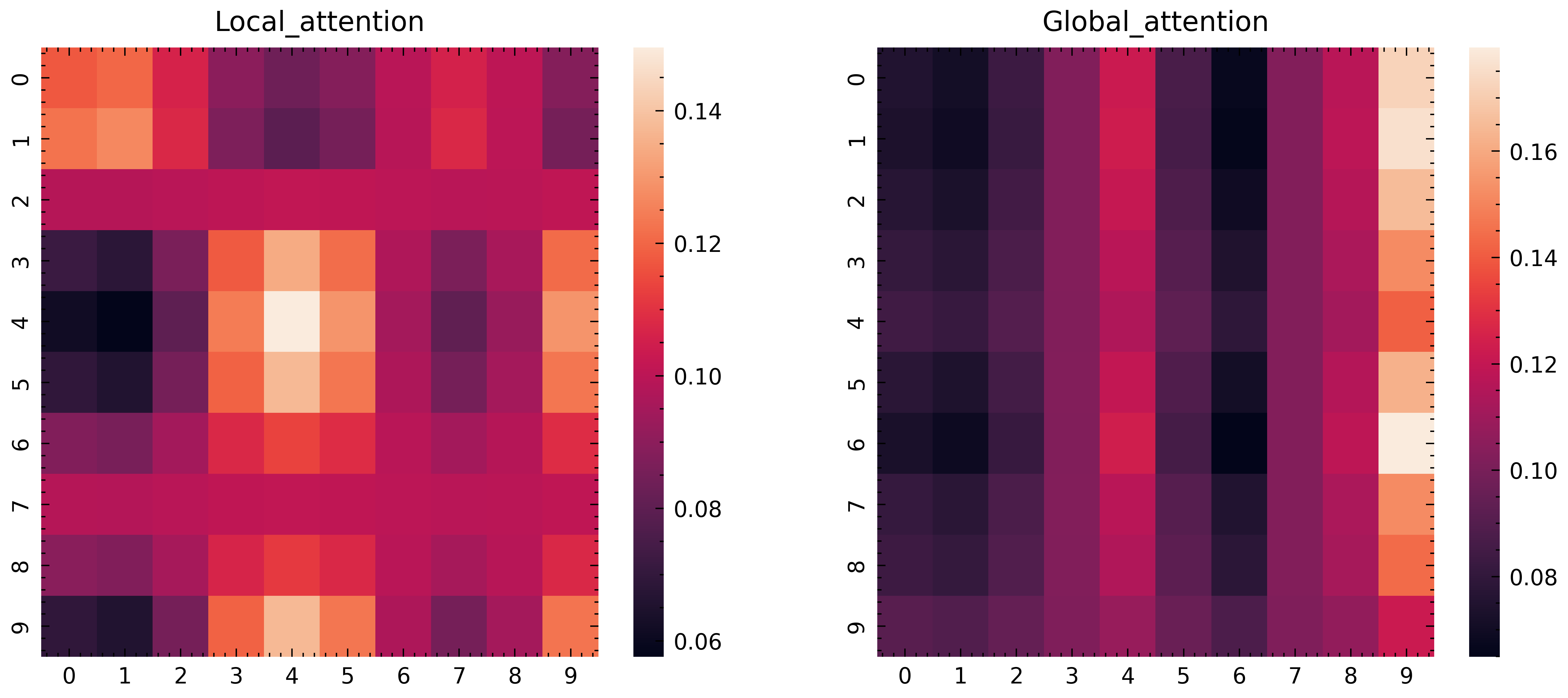}
	\caption{Visualization of the attention matrix.}
	\label{fig:attention matrix}
\end{figure}

We trained our model using the stochastic gradient descent optimizer AdamW~\cite{kingma2014adam} with an initial learning rate of 0.01, and a weight decay of ${10}^{-5}$(using bias correction by default). We also used the step-scheduler with a step size of 5~\cite{saleh2019dynamic} and a learning decay rate of 0.9. For POT parameters, $q={{10}^{-4}}$ for all data sets, low quantile is 0.07 for SMAP, 0.01 for MSL, and 0.001 for others. Note that the number of heads in the multi-head attention mechanism is controlled to be the same size as the dataset size. We use $80\%$ of each time series as the training set and the remaining $20\%$ of data as the testing set. Since this is an unsupervised method, we do not use any labeling information in the training process. For each time series, only the next timestamp is predicted and marked as normal or abnormal. We use the early stopping strategy to train DTAAD, \textit{i.e.}, we stop the training as soon as the verification accuracy starts to drop, this is done to prevent overfitting and reduce the performance fluctuations on the validation set. We incorporate the moving average of validation performance to smooth fluctuations and make more informed decisions about the optimal stopping point. The objective of considering trends rather than single fluctuations is to ensure a more stable convergence of the model. We use the following hyperparameter values.
\begin{itemize}
	\item Window size = 10
	\item Loss weights generated by the two decoders = 0.8
	\item Convolutional kernel size of TCN = 3 \& 4
	\item Number of layers in transformer encoders = 1
	\item Number of layers of encoders feedforward units = 2
	\item Hidden units in encoder layers = 16
	\item Dropout in encoders = 0.2
\end{itemize}

\section{VISUALIZATION}
The performance of the DTAAD model and its variants with different sliding windows is illustrated as shown in Figure~\ref{fig:window size}. We found that if the window is too small or too large, there are problems with contextual information not being captured and local anomalies potentially hidden in a large number of data points in the window (F1 and AUC scores drop). The window size of 10 provides a reasonable balance between F1 scores and training time, and importantly we also found that when the model adds the Tcn\_Global's design component, the final score rebounded as the window size was increased. Figure~\ref{fig:attention matrix}, from its visualization of the attention matrix, we can see that global attention pays more balanced attention to the overall trend, while local attention will focus on the local anomalous part. 

% use section* for acknowledgment
% \section*{Acknowledgment}

% Can use something like this to put references on a page
% by themselves when using endfloat and the captionsoff option.
\ifCLASSOPTIONcaptionsoff
  \newpage
\fi

% trigger a \newpage just before the given reference
% number - used to balance the columns on the last page
% adjust value as needed - may need to be readjusted if
% the document is modified later
%\IEEEtriggeratref{8}
% The "triggered" command can be changed if desired:
%\IEEEtriggercmd{\enlargethispage{-5in}}

% references section

% can use a bibliography generated by BibTeX as a .bbl file
% BibTeX documentation can be easily obtained at:
% http://mirror.ctan.org/biblio/bibtex/contrib/doc/
% The IEEEtran BibTeX style support page is at:
% http://www.michaelshell.org/tex/ieeetran/bibtex/

%\section*{Acknowledgment}
%We would like to express our sincere thanks to \textit{hrstek Co., Ltd.}., Shanghai, China, %for their support of this work. This work was also supported by a special fund for the %development of information technology in Jinshan District, Shanghai (2021-XXH-11).

\balance
\bibliographystyle{IEEEtran}
% argument is your BibTeX string definitions and bibliography database(s)
\bibliography{references}
%
% <OR> manually copy in the resultant .bbl file
% set second argument of \begin to the number of references
% (used to reserve space for the reference number labels box)
% \begin{thebibliography}{1}

% \end{thebibliography}

% biography section
% 
% If you have an EPS/PDF photo (graphicx package needed) extra braces are
% needed around the contents of the optional argument to biography to prevent
% the LaTeX parser from getting confused when it sees the complicated
% \includegraphics command within an optional argument. (You could create
% your own custom macro containing the \includegraphics command to make things
% simpler here.)
%\begin{IEEEbiography}[{\includegraphics[width=1in,height=1.25in,clip,keepaspectratio]{mshell}}]{Michael Shell}
% or if you just want to reserve a space for a photo:

% \begin{IEEEbiography}{Michael Shell}
% Biography text here.
% \end{IEEEbiography}

% if you will not have a photo at all:
%\begin{IEEEbiographynophoto}{John Doe}
%Biography text here.
%\end{IEEEbiographynophoto}

% insert where needed to balance the two columns on the last page with
% biographies
%\newpage

%\begin{IEEEbiographynophoto}{Jane Doe}
%Biography text here.
%\end{IEEEbiographynophoto}

% You can push biographies down or up by placing
% a \vfill before or after them. The appropriate
% use of \vfill depends on what kind of text is
% on the last page and whether or not the columns
% are being equalized.

%\vfill

% Can be used to pull up biographies so that the bottom of the last one
% is flush with the other column.
%\enlargethispage{-5in}

% that's all folks
\end{document}